%% file: main.tex
\documentclass[conference]{IEEEtran}
\IEEEoverridecommandlockouts
\usepackage{cite}
\usepackage{amsmath,amssymb,amsfonts}
\usepackage{graphicx}
\usepackage{svg}
\usepackage{textcomp}
\usepackage{xcolor}
\usepackage{balance}
\usepackage{url}
\usepackage{comment}
\usepackage{hyperref}

\usepackage{booktabs}
\usepackage{algorithm}
\usepackage{algpseudocode}
\usepackage{subcaption}
\def\BibTeX{{\rm B\kern-.05em{\sc i\kern-.025em b}\kern-.08em
    T\kern-.1667em\lower.7ex\hbox{E}\kern-.125emX}}
\begin{document}

\newcommand{\name}{\textsc{PHOENIX}}

\title{\name{}: Resilient LLM Training with Hot-Swapping via Zero-Overhead Checkpoint}
\author{
\IEEEauthorblockN{
Haotian Xie*,
Junlin Chen*,
Mingkai Zheng,
Lishan Yang,
and Zhao Zhang
}
\IEEEauthorblockA{
\textit{Rutgers University, George Mason University}\\
\{haotian.xie, junlin.chen110, mingkai.zheng, zhao.zhang\}@rutgers.edu,
lishan.yang@gmu.edu
}
}

\maketitle

\begin{abstract}
State-of-the-art large language model (LLM) training takes tens of thousands of graphics processing units (GPUs) for months and encounters failures across the software and hardware stack.
Existing fault-tolerance mechanisms either impose non-trivial overhead during failure-free execution or suffer from prolonged recovery latency, particularly under scenarios where a small subset of compute nodes experience permanent failures.
%The tradeoff between failure-free overhead and recovery latency forms a space forms a Pareto frontier
We present \name{} to simultaneously address both optimization objectives. 
\name{} incorporates a fault-tolerance mechanism that restores LLM training via hot-swapping, namely by replacing failed nodes with spare nodes without terminating the complete job.
The hot-swapping of \name{} is enabled by two ideas:
First, it exploits an off-critical-path in-memory checkpointing mechanism for spatial redundancy.
Second, it introduces a communicator reconstruction protocol that replaces failed nodes with spare nodes at runtime.
\name{} efficiently overlaps the in-memory checkpointing with computation, thus introducing zero overhead during error-free execution. 
Upon permanent node failures, \name{} can rebuild memory states with minimal recomputation by leveraging in-memory checkpoints. 
We evaluate \name{} across scales (up to 512 NVIDIA A100 GPUs) and LLMs (up to 65B parameters), and observe zero checkpoint overhead with hot-swapping recovery completing in under 40 seconds.
These results show that \name{} simultaneously achieves both zero-overhead error-free execution and extremely low recovery cost.
\end{abstract}


\input{01intro}

\input{02background}

\input{03related}

\input{04cost}

\input{05design}
\input{06impl}

\input{07expr}

\input{08conclusion}

\balance
\bibliographystyle{IEEEtran}
\bibliography{refs}

\end{document}

%% file: 01intro.tex
\section{Introduction}
LLM training runs on thousands to tens of thousands of GPUs, where node failures are no longer rare events but an expected operating condition~\cite{grattafiori2024llama, opt175logbook}. 
Published interruption statistics from a 32K-GPU LLM pre-training deployment report 678 unexpected interruptions, with GPU HBM memory faults, PCIe device failures, and NCCL watchdog timeouts alone accounting for 49.9\% of all events~\cite{salpekar2026training}. 
The traditional checkpoint-restart solution periodically writes a persistent checkpoint, e.g., for every 1,000 steps~\cite{megatronlm, grattafiori2024llama}, to amortize the checkpointing overhead.
Upon failures, the job terminates, and the user needs to start a new job that reloads the latest checkpoint. 
This approach pays for the recovery cost twice: 
It introduces backup overhead during normal execution, and when a failure occurs, it incurs both recovery latency and the recomputation cost for work performed after the latest checkpoint.
On over-committed supercomputers, such as NERSC Perlmutter or TACC Vista, a double-bit memory error or an NCCL error can cause the complete job to quit.
The user must re-enter an already congested queue, incurring additional wait time.
%Given frequent failures, the goal of a fault-tolerant LLM training framework is to balance the overhead of error-free execution and the recovery cost.
\begin{figure}
    \centering
    \includegraphics[width=1\linewidth]{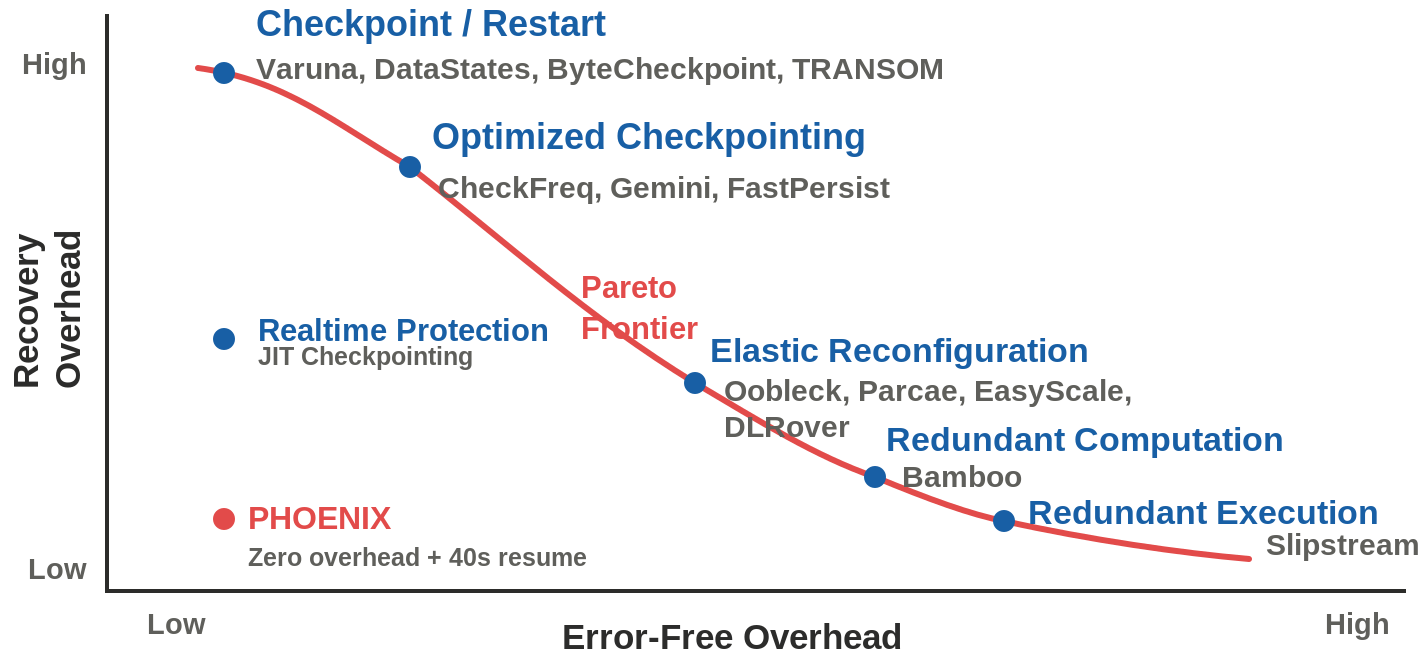}
    \caption{Conceptual landscape of existing fault-tolerance strategies in the tradeoff space of error-free overhead and recovery cost}
    \label{fig:pareto}
    \vspace{-3ex}
\end{figure}

\autoref{fig:pareto} conceptually visualizes existing fault-tolerant LLM training strategies in a tradeoff space between the error-free overhead and the recovery cost.
Checkpointing strategies, such as Datastates-LLM~\cite{maurya2024datastates}, ByteCheckpoint~\cite{10.5555/3767955.3767985}, TRANSOM~\cite{wu2023transom}, and Universal Checkpointing~\cite{10.5555/3768039.3768129}, substantially improve checkpoint throughput and flexibility.
However, they still recover by restarting training from a persistent checkpoint, which may lead to a non-trivial cost due to recomputation for the lost training progress. 
Prior in-memory checkpointing strategies~\cite{264850} reduce write latency, but assume the in-memory checkpoint is accessible in the presence of a failure, which is not a valid assumption if the failure is permanent. 
On the other hand, elastic fault-tolerance strategies, such as Bamboo~\cite{thorpe2023bamboo}, tolerate membership changes by exploiting bubbles in pipeline-parallelism.
But these approaches are with degraded training throughput. 

Our goal in this work is to reduce the overhead for both error-free execution and recovery.
For error-free execution, the LLM framework needs to persist checkpoints asynchronously to avoid potential contention with the training processes.
To lower recovery cost, the LLM framework needs to checkpoint at every iteration to minimize recomputation in the presence of a failure.
Further, the LLM framework needs to avoid a full restart, which introduces additional overhead for global job termination, checkpoint consistency verification, filesystem I/O, and memory state reconstruction.
These challenging requirements motivate the design of \name{}, a fault-tolerant system for LLM training via hot-swapping that leverages zero-overhead asynchronous in-memory checkpointing in complex parallelism.

\name{} is designed with two unique insights.
The first insight is that the memory states during LLM training are distributed using different strategies in complex parallelism, e.g., 3D parallelism.
A typical setting of 3D parallelism, which overlays sharded data-parallelism with pipeline-parallelism and tensor-parallelism, replicates the model shards across data-parallel peers while distributing the optimizer states (i.e., the first and second moments) across GPUs within a data-parallel rank (i.e., a group of GPUs). 
Upon a permanent node failure, model parameters remain reconstructible from healthy ranks that share the same logical partition, but the optimizer state shards cannot be recovered.
Thus, we design a low-overhead, across-nodes in-memory optimizer state checkpoint strategy that effectively overlaps communication with computation, so that \name{} can replicate optimizer state shards in every iteration while avoiding exposed checkpoint overhead. 
A second insight is that \name{} views node failure recovery as an online topology-repair problem rather than a full-job restart.
Thus, we design a distributed communicator reconstruction protocol to replace failed nodes with spare nodes to avoid full allocation restart.

We implement \name{} on top of PyTorch~\cite{paszke2019pytorch} and Megatron-LM~\cite{megatronlm}, but the design relies only on generic capabilities: sharded model and optimizer states, host-side spare-node management, shared recovery coordination, and runtime group reconstruction. 
\name{} includes an error-classification layer that distinguishes transient and permanent failures, allowing \name{} to apply topology-changing recovery for failures that actually invalidate node membership. 
% \zhao{We use both Perlmutter and Vista, right? Say it here.}
We evaluate \name{} on two production HPC systems, i.e., NERSC Perlmutter and TACC Vista, using up to 512 A100 GPUs and 64 GH200 GPUs, respectively.
We inject node failures into live training jobs with reserved spare nodes, which is a standard methodology in research on fault tolerance and recovery~\cite{10.1145/2063384.2063427}.
% ~\zhao{add a citation} 
Across both platforms, we observe negligible per-step  checkpointing overhead on the training critical path.
Moreover, \name{} consistently completes node replacement and restores training around 40\,s after a failure.

%We will release the \name{} prototype, the failure-injection harness, the plotting scripts, and the evaluation artifacts upon acceptance.

Our main contributions are as follows:
\begin{itemize}
    \item We present \name{}, an online hot-swapping recovery mechanism for large-scale LLM training. %It replaces a failed node with a spare node and resumes training from the latest protected in-memory state instead of restarting the entire job.
    \item We design an asynchronous neighbor-memory replication strategy for optimizer state shards. 
    %that enables per-step in-memory checkpointing without introducing additional overhead on the training critical path.
    %For a 2.3B model with TP=1, PP=4, and DP=2, we observe that enabling optimizer-state protection does not incur any measurable overhead, remaining within the natural runtime variance of the baseline training.
    %\item We implement \name{}'s hot-swapping and communication-reconstruction path in a live distributed training stack.
    %On a fixed 20B GPT model with constant global batch size, recovery latency increases only gradually as the system scales from 32 to 512 GPUs, from 18.64\,s to 32.63\,s.
    %Across model scales on a fixed 128-GPU (32-node) cluster, recovery latency remains stable, ranging from 20.72\,s to 22.79\,s for models from 0.20B to 2.3B parameters.
    %\item We show that \name{}'s scaling bottleneck is not parameter restoration but optimizer-state retrieval and the final synchronization needed to resume training, thereby identifying the next optimization target for host-swapping recovery at larger scale.
    \item We evaluate \name{} on up to 512 GPUs and 65B model size, showing zero-overhead during error-free execution and a consistent recovery cost within 40~s. 
\end{itemize}

The zero-overhead checkpointing of \name{} depends on the GPU cluster configuration, the model size, and the model deployment strategies. 
Although \name{} is only examined on Perlmutter and Vista, we expect \name{} to be effective across many other GPU-dense supercomputers. 
This is because the checkpointing overhead is one order of magnitude lower than computation, as shown in \autoref{fig:same_topology_scaling_tp4pp4}, on the current generation machine.
Given the trend of GPU compute capability and interconnect bandwidth increase on the coming NVIDIA Grace-Blackwell supercomputers, \name{} is expected to continue achieving zero-overhead checkpointing.

%% file: 02background.tex
\section{background}
\label{sec:back}

\subsection{Large Language Model Training}

LLM capabilities often increase with model and dataset size~\cite{bommasani2021opportunities,hoffmann2022training}.
%With the ongoing renaissance of deep learning, it is believed that the capability of LLMs increases with larger models and datasets. 
Thus, to overcome training time and memory limitations, researchers have designed various distributed training strategies, including data-, tensor-, pipeline-, and sequence-parallelism~\cite{ben2019demystifying}.

\textbf{Data parallelism} replicates a model across processors.
Training then iterates five steps: 1) I/O, 2) forward compute, 3) gradient evaluation, 4) gradient exchange, and 5) variable update.
\textbf{3D-parallelism} partitions a model across processors by layering data-parallelism over model- and pipeline-parallelism.
Processors are partitioned into groups (\textit{Data Parallel Rank}\/)
%in \autoref{fig:3d-para}), 
and each group hosts a complete copy of the model.
Inside each group, the model is spread across nodes by layers to form the pipeline stages.
Inside each node, layers (tensors) are evenly distributed across multiple processors in what is referred to as \textit{model} or \textit{tensor} parallelism.
With the data-parallel approach, the only communications are the broadcast at the start of training and the gradient exchange in each iteration.
In contrast, 3D-parallelism requires intra- and inter-node communication for both forward and backward computation.
\textbf{Long sequence support}, a fourth dimension in LLM training, enables learning from larger contexts. 
Researchers have designed sequence-parallelism with various partitioning and communication strategies over the query, key, and value matrices~\cite{li2022sequence, liu2024blockwise, liu2023ring}.
Among existing methods, ring attention achieves the lowest memory overhead by using an outer loop of query blocks (across GPUs) and inner loops for key-value blocks (within a GPU).
%, as shown in \autoref{fig:ringattention}.
These distributed LLM training strategies complicate the design of fault-tolerant LLM training systems, as they are prone to suboptimal recovery decisions and unnecessary overhead in the error-free phase.

\subsection{Errors in Systems}

% 1) Errors are common -- show some papers with error rates;
% 2) Source of errors. 
% 3) Common ways to predict failure nodes.
% Software reliability:
% 1) Fault injection study is common in assessing application resilience;
% 2) Protection mechanisms, from single node to distributed systems.

% \lishan{I revised this paragraph to clear the ``fog'' of fault/error/failure}
% \begin{figure}
%     \centering
%     \includegraphics[width=.5\linewidth]{figures/error-faults.png}
%     \caption{Example illustrating faults, errors, and failures.}    
%     \label{fig:error-fault-concept}
% \end{figure}
In computer systems, hardware transient faults (i.e., soft errors)~\cite{nie2016large,beigi2023systematic} can be triggered by high energy particles from cosmic radiation~\cite{8416467}, shrinking transistors, and low voltage operation~\cite{killi}.
These hardware faults manifest as bit-flips and can range from a single bit to multiple bits and even to hundreds of lines of memory cells~\cite{beigi2023systematic}.
% Multi-bit faults have been confirmed in multiple studies~\cite{beigi2023systematic}, sometimes reaching 45\% of total observed faults.
We explain the concepts of fault/error/failure~\cite{snir2014addressing} next:
A hardware \underline{\it fault} propagates in the system; once it reaches the software level, it becomes an \underline{\it error} that 
is visible to the program.
That error further propagates in the program execution, which can lead to three possible outcomes~\cite{de2017radiation}: 
1) no effect (i.e., correct output), 2) silent data corruption, 3) \underline{\it failures} including application crashes and unresponsive systems. 
The first two types of outcome are not noticeable to users, although as quantified in prior research~\cite{zhang2019quantifying, he2023understanding}, silent data corruption can alter the model parameters produced by training.
For noticeable errors, scientists have designed a comprehensive set of technologies for detecting and tolerating the errors. 
As our choices of recovery methods will vary greatly with error type, 
error characterization and analysis are important for understanding and, more importantly, for detecting and correcting errors efficiently~\cite{pinciroli2021lifespan}.
Accurate understanding of errors, and, if possible, accurate prediction of imminent errors and failures, are important if we are to perform timely mitigation actions that
avoid data loss and increases system dependability~\cite{pinciroli2021lifespan}.

\begin{table*}[t]
\renewcommand{\arraystretch}{0.85}
\caption{
Comparison of failure recovery approaches.
\name{} maintains continuously recoverable state and enables online recovery via topology repair.
}
\centering
\small
\begin{tabular}{lccc}
\toprule
\textbf{Method} & \textbf{Checkpoint Timing} & \textbf{Recovery Mechanism} & \textbf{Replay} \\
\midrule
Checkpoint-Restart & Periodic & Restart + reload & Yes (interval) \\
JIT Checkpointing & On failure & Restart + reconstruct state & $\leq$ 1 minibatch \\
Bamboo & None (redundant compute) & Continue from redundant work & None \\
\name{} & Per-step (in-memory) & Node replacement + reconfiguration & $\leq$ 1 step \\
\bottomrule
\end{tabular}
\label{tab:comparison}
\end{table*}

% Error/failure prediction may be formalized as a binary classification problem: to predict, given node  characteristics, the possibility of failure in a given time window.
% %based on the training dataset (usually the server error log), given the  characteristics of the node, to predict the possibility of failure occurence in the given time window.
% A predictor may be constructed by training a statistical model such as Random Forest or XGBoost on server error logs or other datasets~\cite{pinciroli2021lifespan}.
% Based on model predictions, system administrators can investigate problematic nodes and take action accordingly.
% However, this area is still not well explored, and challenges remain:
% 1) \emph{Data imbalance}\/: System logs predominantly contain normal operational behavior, with actual failures being rare. This imbalance poses a significant challenge in accurately identifying potential failures.
% 2) \emph{False positives}\/: While avoiding false negatives is critical, an excessive number of false positives can lead to  unnecessary maintenance actions, introducing unwanted operational overhead and additional costs of addressing these false alarms.

%% file: 03related.tex
\section{Related Works}

\label{sec:related}
Existing fault tolerance solutions can be categorized into two approaches. 
\emph{Checkpointing} methods recover model states from recent checkpoints, while \emph{Exclude-and-Run} methods exclude faulty devices while allowing healthy workers to continue without interruption. 

Early neural network training approaches employed elastic training methods to handle faults, with frameworks like Horovod~\cite{sergeev2018horovod} and PyTorch~\cite{paszke2019pytorch} offering APIs for failure recovery. 
However, these methods primarily address data parallelism for models that can fit on a single GPU. 
As the size of models grows, so has the likelihood of encountering failures during training~\cite{jeon2019analysis,weng2022mlaas,opt175logbook}. EasyScale~\cite{li2023easyscale} and DLRover~\cite{wang2024dlroverrmresourceoptimizationdeep} take hybrid parallelism into account, then propose resilient training pipelines that can be applied to thousands of GPUs. 
Failure rates are even higher when using spot training~\cite{athlur2022varuna,thorpe2023bamboo} in clusters. Varuna~\cite{athlur2022varuna} uses periodic checkpoints to recover from one or more failures, while Bamboo~\cite{thorpe2023bamboo} takes advantage of pipeline parallelism by utilizing bubbles between neighboring stages to create redundancy. 
However, both methods introduce additional re-computing or checkpointing overhead, which can significantly impact training efficiency as failure rates increase. 

Oobleck~\cite{jang2023oobleck} achieves a balance between the throughput and fault tolerance by introducing pipeline templates that enable rapid failure recovery and full GPU utilization. 
Parcae~\cite{duan2024parcae} leverages a novel metric, liveput, to dynamically reconfigure the parallelization strategy in preemptive environments, maximizing throughput. 
Since frequent adjustments to hybrid parallelism can lead to overhead, Slipstream~\cite{gandhi2024slipstream} exploits healthy GPUs to take over computations from failed GPUs in a pipeline stage without changing the strategy. 

Due to the high overhead associated with checkpoint-based recovery, Checkfreq~\cite{mohan2021checkfreq} introduces a fine-grained, automated checkpointing mechanism that overlaps computation with saving model states, optimizing checkpoint frequency through runtime profiling. 
Gemini~\cite{wang2023gemini} leverages hierarchical memory to create a more efficient checkpointing system, while FastPersist~\cite{wang2024fastpersist}, developed by DeepSpeed, optimizes checkpoint writing from NVMe to SSDs in parallel, significantly improving checkpointing efficiency.  
In fault-tolerant scenarios, Just-in-time checkpoint~\cite{gupta2024just} uses data-parallel replicas in 3D parallelism and leverages the iterative nature of DNN training to implement a mechanism that allows at most one step of re-computation in case of failure. 
Techniques~\cite{zhang2023efficient, eisenman2022check, maeng2021understanding} that consider model-specific characteristics (e.g., recommendation models that do not update all parameters simultaneously) further enhance fault recovery efficiency. 
For example, Check-N-Run~\cite{eisenman2022check} copies only modified parameters to host memory, and CPR~\cite{maeng2021understanding} recovers only the failed GPU using checkpoints while allowing other GPUs to continue working.

In general, checkpoint-based methods ensure that the model returns to a globally stable state after a failure, but may introduce additional synchronization overhead that lowers training efficiency, particularly when failure rates are high. Exclude-and-run methods avoid this overhead by dynamically adjusting the parallelization strategy or redistributing computations to functioning workers, but the imbalanced distribution of a mini-batch across the reduced number of GPUs may lead to straggling processes. Existing in-memory approaches such as JIT checkpointing can avoid checkpoint-interval replay, but their reliance on failure-time state reconstruction through function overloading can introduce nested recovery logic and unstable behavior under complex failure scenarios~\cite{10.1145/3581784.3607041}. As shown in ~\autoref{tab:comparison}, \name{} differs fundamentally from all three categories: rather than constructing checkpoints at failure time, relying on redundant computation, or reducing the worker set. \name{} maintains a continuously recoverable state in distributed memory and performs recovery through online node replacement and communication reconfiguration, eliminating full-job restart and reducing recovery to a bounded topology repair process.

%% file: 04cost.tex
\section{Disruption Cost Modeling}
\label{sec:cost}

We model the disruption caused by failures as the sum of three terms:
\begin{enumerate}
    \renewcommand{\labelenumi}{(\theenumi)}
    \item The overhead paid during failure-free execution to maintain recoverability.
    \item The latency required to restore execution after a failure.
    \item The amount of work that must be replayed after recovery.
\end{enumerate}

For periodic checkpoint-restart, if checkpoints are taken every $K$ steps and each step takes $t_{\text{step}}$, then the replay cost of a failure is $R \cdot t_{\text{step}}$, where $R$ is the number of steps executed since the last checkpoint.
Under a uniform-failure approximation, $\mathbb{E}[R] \approx (K-1)/2$.
However, failures in large-scale training are not necessarily uniform; they may be bursty and system-dependent.
We therefore also consider an empirical replay distribution derived from failure traces, replacing the uniform approximation with $\mathbb{E}[R] = \sum_r r \Pr(R=r)$.

This yields the expected disruption per failure under checkpoint-restart:
\[
\mathbb{E}[C_{\text{ckpt}}]
=
C_{\text{overhead}}^{\text{ckpt}}
+
\mathbb{E}[T_{\text{restart}}]
+
t_{\text{step}} \cdot \mathbb{E}[R].
\]

We treat replay as wasted work relative to failure-free execution, even though it is re-executed as part of resumed training.

To relate per-failure disruption to overall training efficiency, we incorporate the system-level failure rate using the mean time to failure (MTTF).
Let $\mathrm{MTTF}_{sys}$ denote the average time between failures at the system scale, and $\lambda = 1/\mathrm{MTTF}_{sys}$ the corresponding failure rate.
The expected disruption per unit time can then be approximated as:

\[
\mathbb{E}[C_{\text{rate}}] = \lambda \cdot \mathbb{E}[C_{\text{ckpt}}].
\]

We further approximate training as alternating between failure-free execution intervals of length $\mathrm{MTTF}_{sys}$ and disruption intervals of expected duration $\mathbb{E}[C_{\text{ckpt}}]$, yielding the effective training efficiency:

\[
\eta \approx \frac{\mathrm{MTTF}_{sys}}{\mathrm{MTTF}_{sys} + \mathbb{E}[C_{\text{ckpt}}]}.
\]

As system scale increases, $\mathrm{MTTF}_{sys}$ typically decreases due to the growing number of failure-prone components, amplifying the impact of recovery cost.

FT-HSDP reports that at O(100K) GPUs, failures occur roughly once every 18 minutes, synchronous recovery stalls the entire job for about 10 minutes, and persistent checkpoints are commonly written every 100 steps, with a typical step time of about 20 seconds.
Under the uniform model, this corresponds to an expected replay cost of approximately $(K/2)\cdot t_{\text{step}} \approx 16.7$ minutes, which is comparable to or even exceeds the restart stall itself.
This highlights that, at a large scale, replay and restart  lead to low overall training efficiency.

\name{} reduces disruption per failure by eliminating checkpoint-interval replay and replacing full-job restart with fast hot-swapping recovery.
The resulting disruption can be expressed as:
\vspace{-1ex}
\[
\mathbb{E}[C_{\text{\name{}}}]
=
C_{\text{overhead}}^{\text{\name{}}}
+
\mathbb{E}[T_{\text{hotswap}}]
+
t_{\text{step}} \cdot \mathbb{E}[\phi],
\]

where $\phi \in [0,1)$ denotes the fractional progress of the interrupted step.
This formulation captures that \name{} reduces replay to at most the interrupted step and avoids full-job restart, significantly lowering the disruption per failure.

%% file: 05design.tex
\section{System Design}
In this section, we describe how \name{} overlaps state protection with training execution and ensures fast, correct recovery without interrupting the global training workflow.

\subsection{Overview}
\label{subsec:overview}
\begin{figure}
    \centering
    \includegraphics[width=1\linewidth, height=5cm]{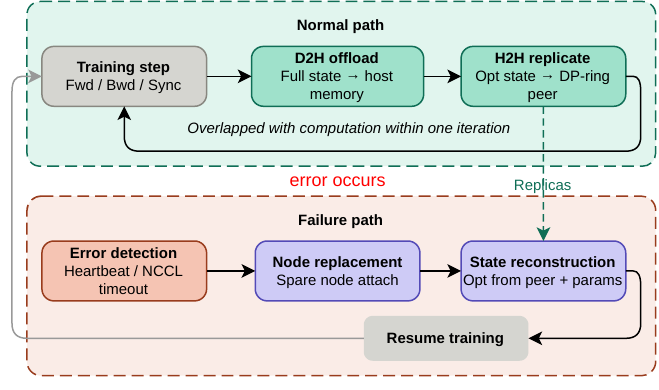}
    \caption{System overview of \name{}. The normal path (top) runs a two-phase offload pipeline overlapped with each training step. The failure path (bottom) detects faults, replaces the failed node, and reconstructs state from in-memory replicas.}
    \label{fig:workflow}
\end{figure}
\begin{figure}
    \centering
    \includegraphics[width=\linewidth]{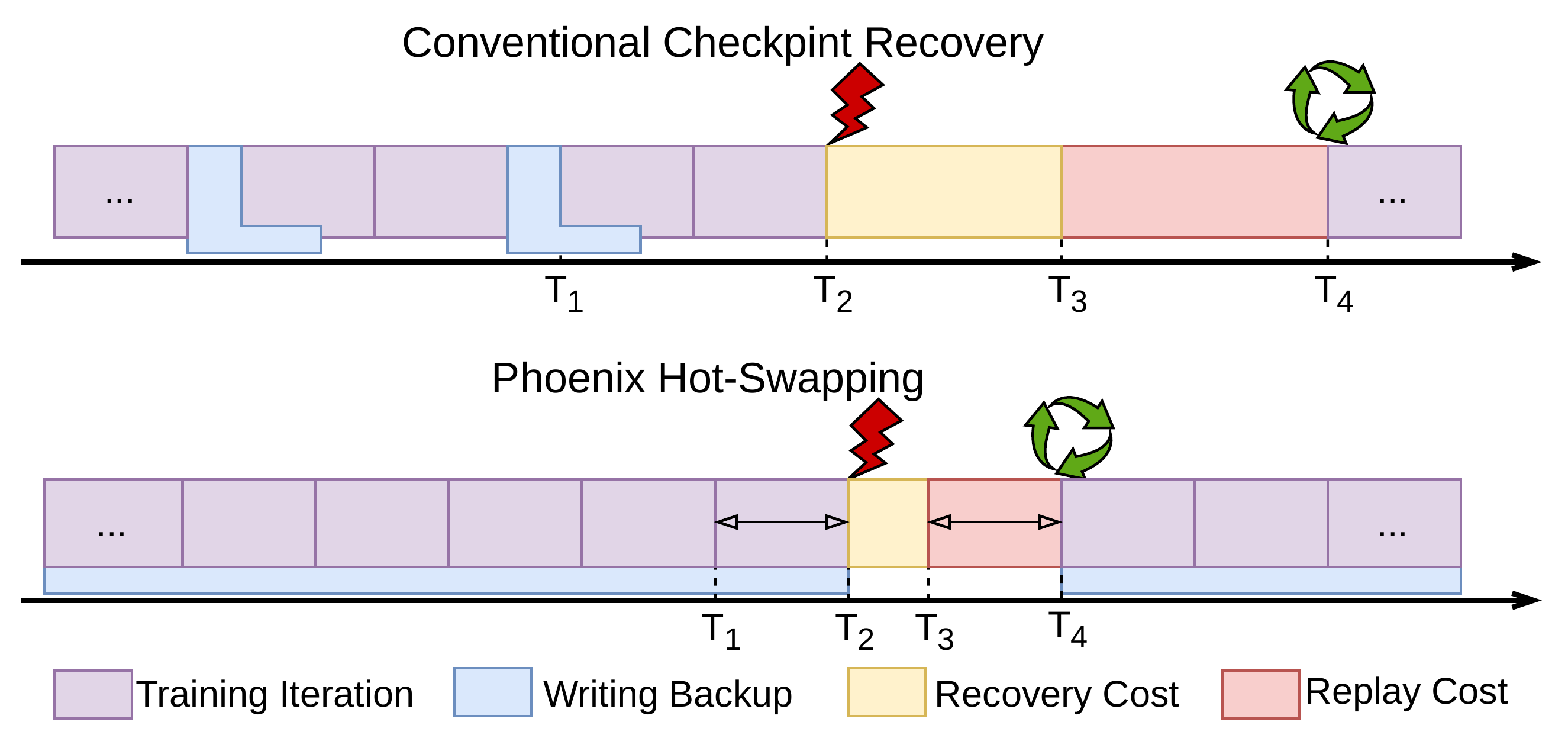}
    \caption{
Timeline comparison of conventional checkpoint recovery and \name{}. 
\name{} overlaps protection with training and reduces disruption after failure.
}
    \label{fig:TimelineComparison}
    \vspace{-2ex}
\end{figure}

As shown in Section~\ref{sec:cost}, failure-induced disruption in large-scale training is dominated by restart latency and replay cost under checkpoint-restart.
This motivates a design that eliminates checkpoint-interval replay and avoids full-job restart.
\name{} achieves this by transforming recovery into an online topology repair problem:
instead of restoring execution from a persistent checkpoint, \name{} maintains a continuously recoverable state in distributed memory and replaces failed nodes with healthy ones, allowing training to resume without restarting the job.

As illustrated in Figure~\ref{fig:workflow}, \name{} comprises three components that separate normal-path state protection from failure-path recovery:
\begin{itemize}
    \item \textbf{Asynchronous Offload Pipeline} (Section~\ref{subsec:async_pipeline}) runs on the normal path once per training iteration. It copies the full rank state to host memory  and replicates optimizer-state shards to a DP-ring peer. Both phases are overlapped with forward, backward, and gradient synchronization, keeping redundancy cost off the critical path.
    \item \textbf{Error Detection and Isolation} (Section~\ref{subsec:error_detection}) monitors node health via heartbeats and NCCL timeouts. When a failure is detected, it traps the exception before it propagates to the job manager, holding surviving ranks at a barrier while a replacement node is provisioned.
    \item \textbf{Recovery Policy} (Section~\ref{subsec:recovery-policy}) executes on the failure path. It attaches a spare node, rebuilds communication groups, and reconstructs the missing state from in-memory replicas. Training resumes from the last completed step with at most one step of re-execution.
\end{itemize}

During normal execution, only the offload pipeline is active, and its cost is hidden within each training iteration.
When a failure occurs, the error detection module triggers the recovery policy, which leverages the remote replicas to restore training without a full job restart.

Figure~\ref{fig:TimelineComparison} compares conventional checkpoint-restart with \name{} from a timeline perspective. 
Let \(T_1\) denote the latest recoverable state before failure.
\(T_2\) is the failure time.
\(T_3\) is the point at which training resumes, and \(T_4\) is the point at which execution catches up to a failure-free trajectory.
In conventional checkpoint-restart, periodic backup operations are placed on the training critical path, and a failure introduces two sources of disruption: explicit recovery latency and replay of lost work between \(T_1\) and \(T_2\). As a result, training cannot resume until the system reloads state and reconstructs execution, and additional time is spent re-executing previously completed iterations before reaching \(T_4\).

In contrast, \name{} decouples state protection from the critical path by continuously maintaining recoverable state in memory and overlapping redundancy operations with normal execution. Upon failure, the system performs online node replacement and reconstructs only the missing state, allowing training to resume near \(T_3\) with at most one step of re-execution. This design significantly reduces both replay cost and end-to-end recovery latency, effectively shrinking the disruption window between failure and full recovery.

\subsection{Asynchronous Offload Pipeline}
\label{subsec:async_pipeline}

\begin{figure}
    \centering
\includegraphics[width=1\linewidth, height=4cm]{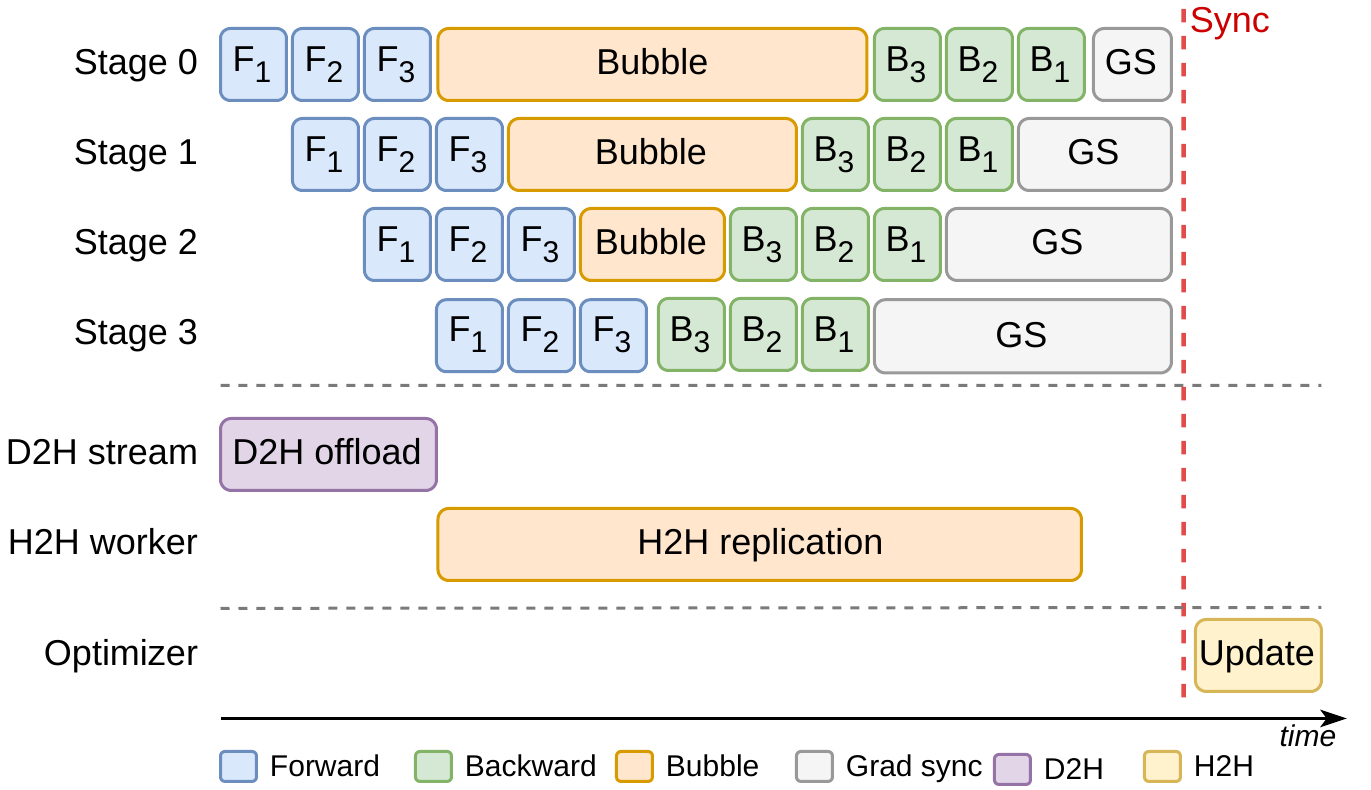}
    \caption{Illustration of the asynchronous offload pipeline across pipeline-parallel stages. 
Computation (forward and backward passes) is interleaved with gradient synchronization, while device-to-host (D2H) offload and host-to-host (H2H) replication progress concurrently on separate execution paths. 
The timeline highlights the overlap between compute and communication throughout an iteration.}
    \label{fig:pipeline}
    \vspace{-2ex}
\end{figure}

The offload pipeline runs once per training iteration,
overlapped with the forward pass, backward pass, and gradient
synchronization of the same iteration as illustrated in \autoref{fig:pipeline}. It proceeds in two phases
whose payloads differ by design.

\paragraph{Device-to-host (D2H)}
The full rank state, which comprises the optimizer-state shard,
the local model-parameter shard, and a metadata envelope
(step index, parallelism configuration, RNG states), is
packed into a contiguous buffer and copied asynchronously
to pinned host memory on a dedicated CUDA stream.
Retaining the complete rank state on the host enables
fast local recovery (e.g., after a transient GPU error)
without any network transfer.

\paragraph{Host-to-host (H2H)}
Only the optimizer-state shard and the metadata envelope are transmitted to the peer node in a ring topology, which is the rank sharing the same tensor-parallel and pipeline-parallel positions but with a different data-parallel index.
Model parameters are excluded: upon node failure, the
replacement rank can reconstruct its parameter shard
from healthy ranks at the same TP/PP coordinate via
a targeted collective, so replicating parameters to a
peer in advance would consume network bandwidth without
reducing recovery latency.
A background worker thread waits on the D2H completion
event and then sends the optimizer payload to the ring
peer via chunked MPI over the DP-local communicator.

This payload is asymmetric, full rank state for D2H versus optimizer-only for H2H, because of the bandwidth difference of the system. Local PCIe transfers are fast, whereas inter-node network bandwidth is more constrained. By transmitting only the optimizer shard over the network, the H2H phase reduces its transfer volume, increasing the likelihood that both phases complete within a single training iteration.

\subsection{Recovery Policy and Correctness Guarantees}
\label{subsec:recovery-policy}

\name{} handles failures by determining whether the affected node can continue participating in distributed execution.

\textbf{Failure Model.}
We consider two classes of failures. 
(1) \emph{Transient failures}, where the affected node remains operational and its state remains valid. 
(2) \emph{Permanent failures}, where a node or its state is permanently lost and can no longer contribute its assigned shards. 
In the latter case, the system replaces the failed node with a spare node and reconstructs the missing state.

Failures that may compromise training semantics, such as software bugs, numerical errors, or silent data corruption, are treated as outside the scope of \name{} and handled by an external fallback mechanism.

\textbf{System Invariants.}
To ensure correctness under failures, \name{} maintains the following invariants:

\begin{itemize}
    \item \textbf{I1 (Topology Consistency).} All involved nodes agree on a single communication topology and shard placement.
    \item \textbf{I2 (Shard Completeness).} For every logical shard, there exists exactly one valid owner among the involved nodes.
    \item \textbf{I3 (Optimizer Availability).} The optimizer state for every shard is available from in-memory replicas.
    \item \textbf{I4 (Step Atomicity).} Training progresses in discrete steps; partial progress from a failed step is not committed.
\end{itemize}

During transient failures, the LLM training system preserves all invariants.
These failures can be recovered in place without modifying  involved nodes or the communication topology.

\textbf{Recovery Procedure.}
For permanent failures, \name{} recovers in three stages.
First, the system selects a spare node and updates the set of involved nodes. This change invalidates the pre-failure communication topology.
Second, all involved nodes are quiesced at a safe boundary, and the system re-establishes a consistent communication topology and shard placement across all nodes. This step ensures that Invariants I1 and I2 hold before execution resumes.
Third, the system reconstructs the missing state on the spare node. Model parameters are reconstructed from healthy nodes that share the same partition, while optimizer state is restored from in-memory replicas. This step ensures that Invariant I3 holds.
Finally, execution resumes from the last completed step, ensuring that Invariant I4 is preserved.

\paragraph{Tolerating Multi-Node Failures via Multi-Peer Replication.}
By default, \name{} replicates each optimizer-state shard to one DP-ring neighbor, tolerating any single-node failure. To guard against concurrent multi-node failures, \name{} can be configured to replicate each shard to $k$ distinct DP peers, with the additional H2H cost scaling linearly in $k$ but remaining overlappable with computation for a sufficiently small $k$.

A shard is irrecoverably lost only if its owner and all $k$ replica holders fail within the same step. We quantify this risk using published failure statistics from Meta's 100K-GPU deployment, which reports one interruption roughly every 18 minutes~\cite{salpekar2026trainingllmsfaulttolerant}. With a step time of approximately 20\,s, this corresponds to a per-node per-step failure probability on the order of $p \approx 10^{-6}$, which is an aggressive estimate that we adopt as an upper bound for our analysis.

\textit{Independent failures.}
When replica placement is topology-unaware, we assume node failures are independent. The probability that a specific shard loses all $k{+}1$ copies in a single step is $p^{k+1}$. Applying a union bound over $D$ data-parallel shards, the per-step probability of any shard becoming unrecoverable is:
\begin{equation}
  P_{\text{loss}} \leq D \cdot p^{k+1}.
\end{equation}
With $D=128$ and $k=1$, this yields $P_{\text{loss}} \leq 1.28 \times 10^{-10}$ per step, or a cumulative probability of $1.28 \times 10^{-5}$ over a $10^5$-step run---fewer than one event per 100 full training runs. Setting $k=2$ reduces the cumulative probability to $1.28 \times 10^{-11}$, which is effectively zero.

\textit{Correlated failures with topology-aware placement.}
The independent-failure analysis breaks down when a shared infrastructure component---such as a top-of-rack (ToR) switch, a network cable, or a power distribution unit---fails and takes down multiple co-located nodes simultaneously. If a shard's owner and its replica reside under the same switch, a single switch failure can destroy both copies.

To mitigate this, \name{} can leverage cluster topology information to place the $k$ replicas on nodes under $k$ distinct switches or racks, ensuring that a single infrastructure event affects at most one copy of any given shard. Let $q$ denote the per-domain per-step failure probability (e.g., a ToR switch failure rate). With replicas distributed across independent failure domains, the per-shard loss probability becomes $q^{k+1}$. Even under a relatively high domain failure rate of $q = 10^{-4}$, $k=1$ yields a cumulative risk of $D \cdot q^{2} \times 10^5 = 0.128$ over $10^5$ steps---non-negligible for safety-critical runs. Setting $k=2$ reduces this to $D \cdot q^{3} \times 10^5 = 1.28 \times 10^{-5}$, rendering correlated shard loss negligible.

In practice, $k=1$ with topology-aware placement suffices for the vast majority of production scenarios, while $k=2$ provides an additional safety margin for deployments with frequent infrastructure-level failures.

%% file: 06impl.tex
\section{Implementation}
\label{sec:implementation}

We implement \name{} as a hybrid design that combines a data-plane extension to Megatron-LM with an external control plane. The data plane integrates directly with Megatron-LM’s distributed execution model, including checkpoint metadata, sharded optimizer abstractions, and 3D-parallel process-group organization,  leaving the failure-free execution path unchanged when the mechanism is disabled.

The control plane is implemented as a lightweight standalone service that coordinates failure detection, node replacement, and recovery orchestration across all involved nodes. It provides a persistent coordination layer independent of the training processes, enabling recovery to proceed even when a subset of training nodes fail.

The implementation spans four key components:
\begin{enumerate}
    \renewcommand{\labelenumi}{(\theenumi)}
\item the program entry path for initializing failure-aware execution and registering with the control plane, 
\item the training-step scheduler for enforcing safe quiescence and recovery boundaries, 
\item the checkpoint and optimizer wrapper for maintaining in-memory recoverable state, and
\item an asynchronous MPI-based transport service for state replication and recovery. 
\end{enumerate}

We implement \name{} as an opt-in extension inside Megatron-LM. The in-process design has two advantages over an external wrapper: it reuses Megatron's distributed-optimizer serialization and 3D-parallel process-group layout directly, avoiding a redundant checkpoint stack; and when the mechanism is disabled, no interception layer is present in the training loop, so the failure-free path incurs zero overhead.

% \subsection{Network Transport}
% \label{subsec:network}

% \textbf{Responsibility.}
% Provide a communication substrate for state replication
% that coexists with the training collectives without
% resource contention.

% \textbf{Mechanism.}
% The offload transport requires a communication channel
% independent of the NCCL communicators used for training
% collectives.
% We use MPI for this purpose.
% A key constraint is that both stacks share the same
% set of network interfaces, and initialization order
% determines resource allocation: if NCCL is initialized
% first, it exclusively claims all available VNI contexts,
% leaving no resources for the offload transport.
% \name{} therefore initializes the offload
% transport before Megatron constructs any NCCL
% communicator, ensuring that it secures its transport
% resources first.
% The offload communicators are further scoped to
% per-data-parallel groups, so offload traffic is
% carried on disjoint communicators from training
% collectives and does not contend in steady state.

\subsection{In-Memory State Protection}
\begin{figure}
    \centering
    \includegraphics[width=\linewidth,height=4cm]{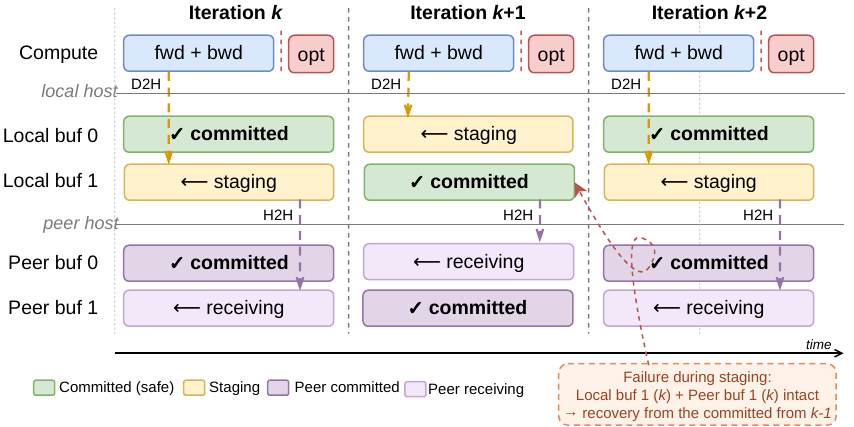}
    \caption{Ping-pong double buffering across iterations. 
Two local buffers alternate between committed and staging states, while peer buffers alternate between committed and receiving. 
Computation overlaps with data movement (D2H and H2H), and buffer roles switch every iteration.}
    \label{fig:double buffer}
\end{figure}
%\textbf{Responsibility}
This component maintains a consistent, immediately recoverable snapshot
of each rank's training state in host memory at all times.
%\textbf{Mechanism}
As shown in ~\autoref{fig:double buffer} offload manager uses a ping-pong scheme for the
local snapshot buffer: two host-side buffers alternate
roles across iterations, one holding the committed
snapshot from iteration $k{-}1$ and the other serving
as the staging target for iteration $k$.
If a failure interrupts the current offload or the
subsequent peer transfer, the in-progress buffer may
be incomplete, but the previous buffer remains intact
and immediately usable for recovery.
This design guarantees that a valid recovery point is
always present in host memory without requiring the
offload to be atomic.
The buffer received from the ring predecessor is
maintained separately and follows the same alternation
logic.

The training loop exposes a single synchronization
point: a wait immediately before the optimizer step
that guarantees the peer-resident replica is fully
committed before the optimizer mutates the protected
state.
In the common case, both D2H and H2H transfers complete
within the iteration's compute window, so this wait
observes an already-finished offload and adds no
exposed stall.

\begin{table}[t]
\label{tab:error}
\renewcommand{\arraystretch}{0.85}
\caption{Error classification and recovery actions in \name{}.}
\label{tab:error_classification}
\centering
\small
\begin{tabular}{ll}
\toprule
\textbf{Failure Category} & \textbf{Action} \\
\midrule
Transient communication faults & Local recovery first \\
GPU memory faults & Node replacement \\
PCIe / host / kernel / reboot faults & Node replacement \\
Network or storage faults & Node replacement \\
Software bugs / Unknown failures & External fallback \\
\bottomrule
\end{tabular}
\end{table}

\subsection{Error Classification}
\label{subsec:error_detection}
Following the failure model defined in Section~\ref{subsec:recovery-policy}, \name{} implements a classification layer that maps observed failures to recovery actions as illustrated in ~\autoref{tab:error_classification}.
The classification is driven by runtime signals that indicate loss of node participation or loss of reliable state. 
Device-level failures (e.g., GPU memory errors), system-level failures (e.g., host crashes or reboots), and persistent communication failures are treated as node-replacement events, as they prevent the node from safely contributing its assigned shards. 
Transient communication failures are initially handled as locally recoverable when the node remains responsive, but are elevated if they lead to sustained communication failure or node unavailability.
When a failure is classified as locally recoverable, \name{} performs a local reset by clearing the node's state and reinitializing it as a replacement node. 
From the system perspective, this is equivalent to a special case of node replacement where the failed node re-enters the system after reset. 
For clarity, we treat both cases uniformly as hot-swapping, and focus the following discussion on node replacement.

To evaluate the system under controlled yet realistic failure conditions, we inject errors during live training runs using timer-based triggers.
At predefined execution points, the timer activates different fault types, including fail-stop events (via process termination), NCCL communication errors and storage-related errors.
This unified injection mechanism enables reproducible evaluation of both abrupt node failures and runtime error conditions within the same framework.

\subsection{Control Plane}
%\textbf{Responsibility}
This component provides a fault-isolated coordination substrate for involved nodes, maintaining globally consistent recovery metadata throughout failure handling and node replacement.
%\textbf{Mechanism}
\name{} implements the control plane as a standalone TCPStore-based service running as an independent process. 
It maintains a key-value store to recover metadata, including failure notifications, recovery epochs, node assignments, and synchronization signals. 
Because the control plane is decoupled from training processes, it remains available when a subset of nodes fails, allowing surviving and replacement nodes to continue coordination. 

The control plane supports dynamic node participation, allowing nodes to join during recovery without requiring global re-initialization. 
Each involved node publishes a recovery descriptor that encodes its logical role, including shard ownership and parallel partition placement. 
These descriptors are used to derive deterministic mappings between source and replacement nodes based on logical shard identity.

\subsection{Recovery Scheduling and Topology Reconfiguration}
%\textbf{Responsibility}
This component orchestrates the recovery process across involved nodes, ensuring consistent execution progression during node replacement and topology reconstruction.
%\textbf{Mechanism}
When a failure requires node replacement, the runtime initiates a recovery epoch and coordinates all involved nodes through the control plane, satisfying I1 mentioned in \ref{subsec:recovery-policy}.
Each node captures its local recovery state in memory and synchronizes before  recovery. 
Nodes are paused at a step boundary to establish a consistent execution point.

A spare node is activated and incorporated into the set of involved nodes. 
The runtime reconstructs communication groups and shard placement based on the updated topology, and all nodes resume execution under the same communication configuration. 
Recovery epochs are serialized, so that a new recovery begins only after the current one completes.

\subsection{State Restoration and Transport}

After topology reconfiguration, \name{} reconstructs the state of failed shards on the spare node. 
Model parameters are restored by transferring shards from source nodes that share the same logical partition, while optimizer state is restored directly from in-memory replicas maintained during training, avoiding access to persistent checkpoints.

To ensure correct reconstruction under dynamic rank reassignment, source and replacement nodes are matched using logical shard identity derived from recovery descriptors rather than post-failure rank indices. 
This guarantees that each shard is reconstructed from the correct source regardless of how ranks are reassigned, corresponding to I2 and I3.
State transfer is implemented by an asynchronous MPI-based transport layer, which allows communication to overlap with recovery preparation on the receiving node and reduces recovery latency.
\vspace{-1ex}
\subsection{Training Resume}

Once model parameters and optimizer state are restored, the spare node assumes the role of the failed node and joins the set of involved nodes, ensuring the I4 completeness.
Training resumes under the reconstructed communication topology from the last completed step, and any partial work from the interrupted step is discarded to preserve consistency. 
If a subsequent failure occurs, \name{} initiates a new recovery epoch following the same procedure.

% \begin{algorithm}[t]
% \caption{\name{} recovery path}
% \label{alg:deadpool_recovery}
% \begin{algorithmic}[1]
% \State classify the failure
% \If{the failure is locally recoverable}
%     \State attempt local recovery
%     \If{local recovery succeeds}
%         \State return to normal training
%     \EndIf
% \EndIf
% \If{the failure is not eligible for node replacement}
%     \State invoke external fallback
%     \State return
% \EndIf
% \State capture the in-memory recovery state
% \State notify healthy ranks and quiesce them
% \State activate the spare node and rebuild the communication topology
% \State restore parameter shards
% \State restore optimizer state
% \State resume training
% \end{algorithmic}
% \end{algorithm}

%% file: 07expr.tex
\section{Experimental Evaluation}

We evaluate \name{} along three dimensions. First, we quantify its checkpoint overhead in the error-free case. Second, we measure the recovery cost of conventional checkpoint/restart under failures. Third, using an error model, we compare the end-to-end time-to-solution of \name{} against checkpoint-based recovery under realistic failure scenarios. Across the configurations we study, \name{} consistently reduces total wall-clock time.

\subsection{Experimental Setup}

\subsubsection{Platforms}
We evaluate \name{} on two production supercomputers: Perlmutter at NERSC and Vista at TACC.

Perlmutter is an HPE Cray EX system. We use its GPU nodes, with each containing one 64-core AMD EPYC 7763 CPU, four NVIDIA A100 GPUs, and 256\,GB of DDR4 host memory. The four GPUs are interconnected via third-generation NVLink. Each node is equipped with four HPE Slingshot 11 NICs, each providing 200\,Gb/s, for an aggregated bandwidth of 800\,Gb/s per node. GPU--CPU and NIC--CPU connectivity are both provided through PCIe 4.0. The Perlmutter interconnect uses a three-hop dragonfly topology.

Vista is an NVIDIA Grace-based supercomputer at TACC. We use Grace--Hopper (GH) nodes, each containing one 72-core NVIDIA Grace CPU and one NVIDIA H200 GPU with 96\,GB of HBM3 memory, connected through the Grace--Hopper superchip interconnect. Each node also provides 116\,GB of LPDDR host memory and connects to the system fabric via NVIDIA NDR InfiniBand at 400\,Gb/s. Vista employs a fat-tree topology.

% \subsubsection{Software Environment}

% On Perlmutter, we use the Cray programming environment with GCC 13.2,
% CUDA 12.4, cuDNN 9.5.0, NCCL 2.24.3, Cray MPICH 8.1.30,
% libfabric 1.22.0, Cray LibSci 25.09.0, and PyTorch 2.6.0-1.
% On Vista, we use the site-provided software environment with NVIDIA 24.7,
% NVPL 24.7, UCC 1.7.0, UCX 1.20.0, Open MPI 5.0.5, and CMake 4.1.1.
\subsubsection{Training Configuration}
We train GPT-style transformer models from 0.6\,B to 65\,B parameters, varying the memory footprint by two orders of magnitude.

Unless otherwise noted, all experiments use full-precision (FP32) training: model parameters, main gradients, and Adam optimizer states (first and second moments, together with the FP32 master copy) are stored in \texttt{torch.float32}, with BF16 and FP16 disabled. This setting maximizes the per-rank state volume and therefore represents a demanding configuration for \name{}'s offload path.
% \begin{table}[t]
% \centering
% \caption{GPT model configurations used in the evaluation. Parameter counts count input and output embeddings separately.}
% \label{tab:model-sizes}
% \small
% \begin{tabular}{lrrrr}
% \toprule
% \textbf{Params} & \textbf{Layers} & \textbf{Hidden} & \textbf{Heads} \\
% \midrule
% 0.61\,B & 16 & 1536 & 24 \\
% 1.41\,B & 24 & 2048 & 32 \\
% 2.30\,B & 32 & 2304 & 36 \\
% 7\,B    & 32 & 4096 & 32 \\
% 21\,B   & 60 & 5376 & 48 \\
% 65\,B   & 80 & 8192 & 64 \\
% \bottomrule
% \end{tabular}
% \end{table}

We use Megatron-LM with 3D parallelism and ZeRO-2 distributed optimizer. Unless otherwise noted, all experiments use the 1F1B pipeline schedule. We evaluate \name{} across multiple parallel configurations from 8 to 512 GPUs.

\subsection{\text{\name{}}'s performance} 
\label{sec:experiment}

We design our evaluation to answer two key questions about \name{}:
\begin{enumerate}
    \renewcommand{\labelenumi}{(\theenumi)}
\item  Does per-iteration in-memory checkpointing introduce any runtime overhead under scaling?
\item  Does hot-swapping recovery remain efficient as system scale increases?
\end{enumerate}

To this end, we conduct two sets of experiments. Firstly, we evaluate the runtime overhead of \name{} under weak scaling, examining whether enabling per-iteration checkpointing affects training step time as model size and cluster size grow. Secondly, we measure the recovery latency of hot-swapping across different cluster scales to characterize its scalability and identify dominant cost components.

\subsubsection{Per-iteration Checkpoint Overhead under Scaling}

\begin{figure}[t]
    \centering
    \includegraphics[width=\linewidth, height=4cm]{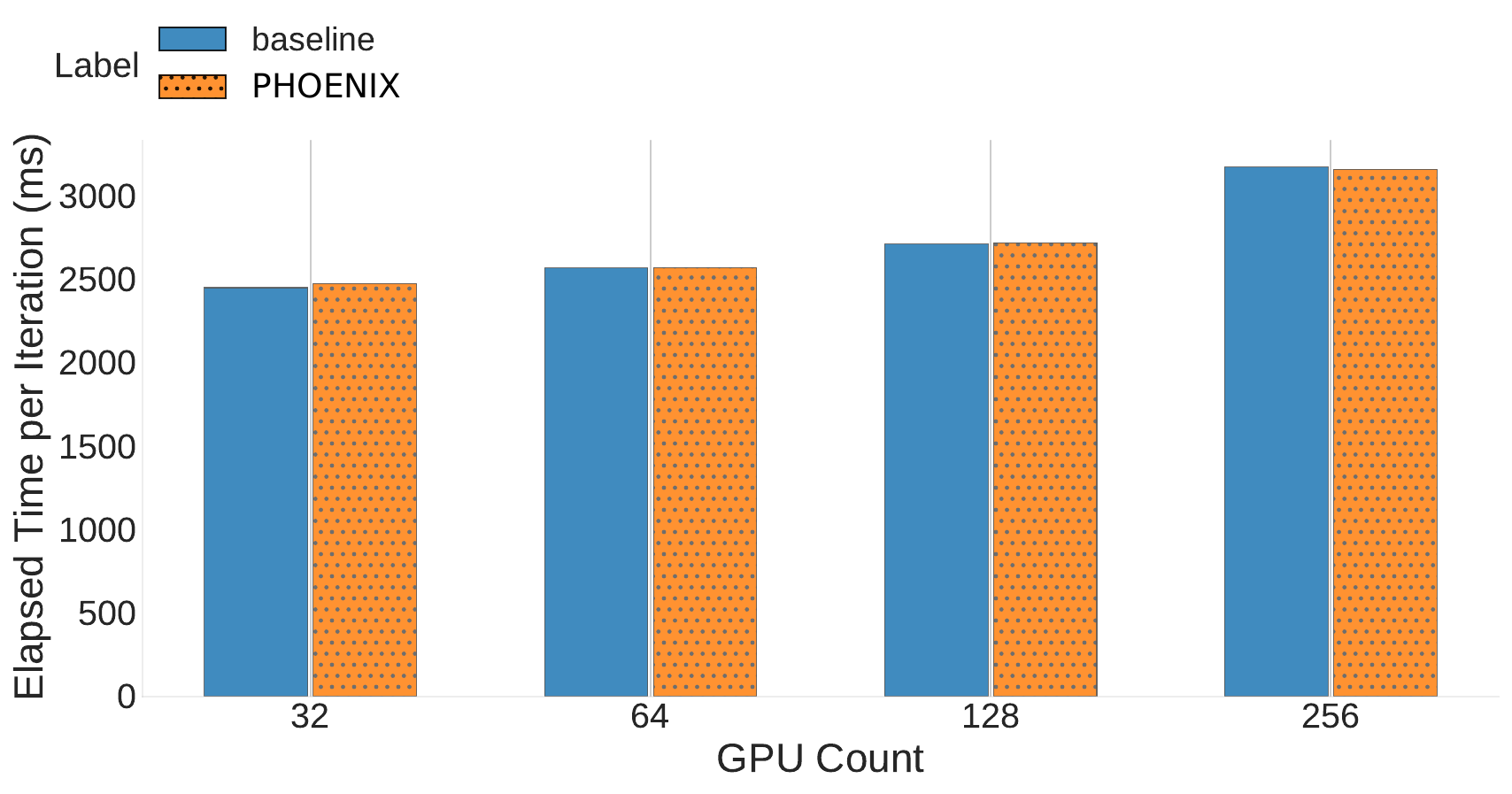}
    \caption[Same-topology scaling (TP=4, PP=4)]%
    {Same-topology weak scaling results for the 2.3B model with tensor parallelism (TP=4) and pipeline parallelism (PP=4). 
    The x-axis shows the number of nodes (8, 16, 32, and 64), and the y-axis reports the per-iteration training time in milliseconds. 
    The blue bars (\textit{baseline}) correspond to the original training configuration without \name{}, while the orange dotted bars (\name{}) show performance with \name{} enabled.}
    \label{fig:same_topology_scaling_tp4pp4}
\end{figure}

We first evaluate whether \name{} introduces any checkpoint overhead in the error-free case. \autoref{fig:same_topology_scaling_tp4pp4} shows weak scaling for the 2.3B GPT model under a fixed parallel topology, with TP=4 and PP=4 held constant. 
As the system scales from 32 to 256 GPUs, we increase the global batch size proportionally to the data-parallel degree, keeping the per-rank workload unchanged. 
To report per-iteration measurement, we discard the first 10 iterations as warm-up, remove the slowest five percent of the remaining iterations, and average the rest. 
Across all scales, \name{} closely tracks the baseline. 
On eight nodes, the baseline averages 2442.13~ms while \name{} averages 2470.15~ms.
On 16 nodes, the two are nearly identical (2560.35~ms vs.\ 2561.75~ms). On 32 and 64 nodes, \name{} is marginally faster (2696.62~ms vs.\ 2701.87~ms and 3122.86~ms vs.\ 3134.81~ms, respectively). 
The differences are small and non-monotonic, with no trend of increasing overhead as the system grows, indicating that \name{} introduces negligible runtime overhead during error-free execution.

\begin{figure}
    \centering
    \includegraphics[width=1\linewidth]{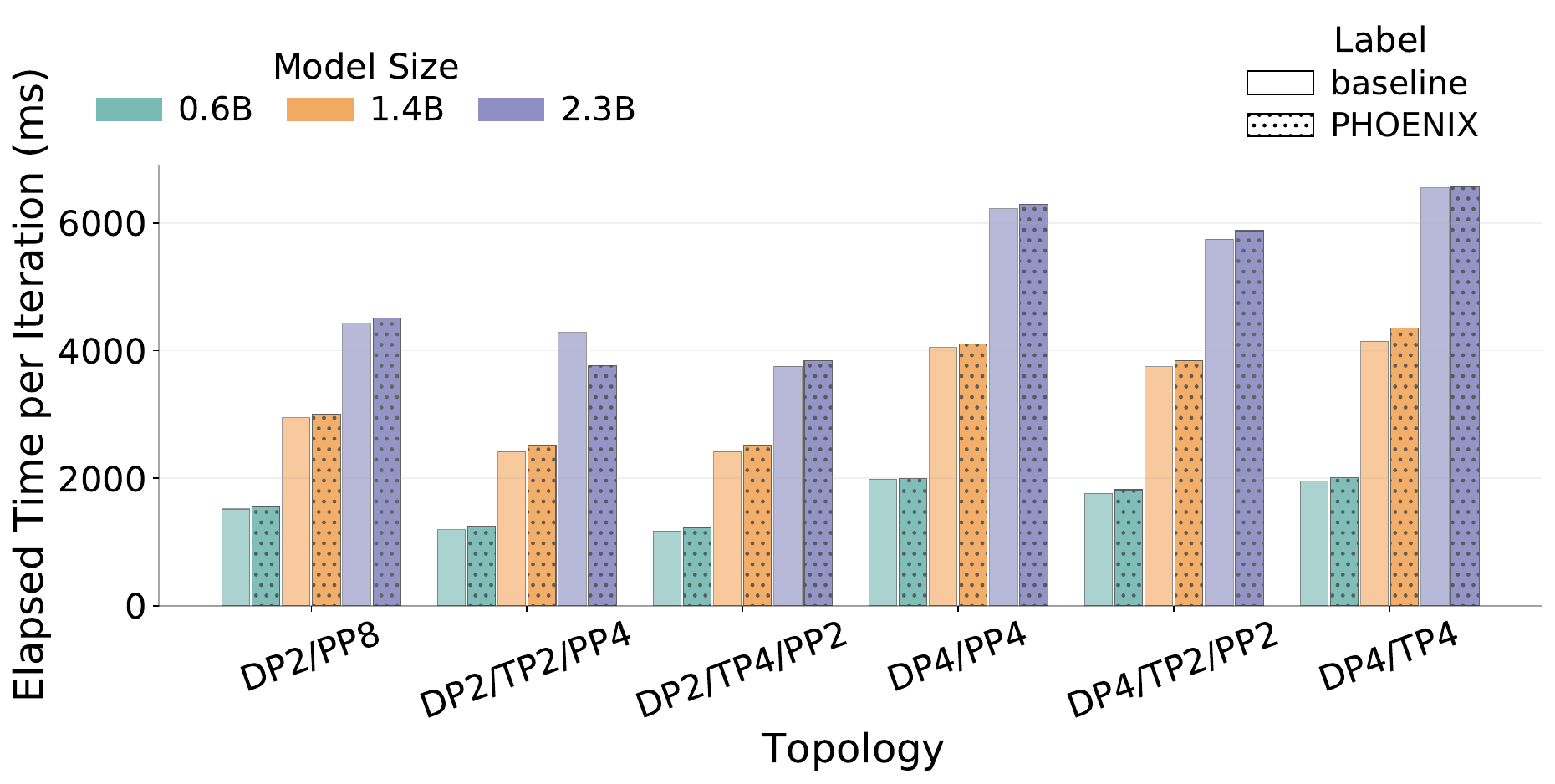}
    \caption{Per-iteration training time across six 3D parallel topologies on 4 nodes (16 GPUs) under weak-scaling settings. The x-axis lists the training topology, and the y-axis shows elapsed time per iteration in milliseconds. Colors denote model size. For each model size within a topology, the solid bar represents the baseline configuration and the hatched bar represents \name{}.}
    \label{fig:topology_sweep_4nodes}
    \vspace{-4ex}
\end{figure}
In addition to the fixed-topology weak-scaling results above, we also evaluate \name{} across a range of 3D parallel layouts on 16 GPUs and 8 GPUs. 
\autoref{fig:topology_sweep_4nodes} compares the baseline and \name{} across six topologies and four model sizes. 
Across all configurations, \name{} closely tracks the baseline. 
Although a few cases show small positive or negative deviations, these differences are non-systematic and are consistent with normal run-to-run and node-allocation variability, rather than a persistent overhead introduced by per-iteration checkpointing. 
Together with the same-topology weak-scaling results, these results show that \name{} introduces no observable checkpoint overhead both as system scale increases and across diverse parallel topologies.
\begin{figure}
    \centering
    \includegraphics[width=1\linewidth, height=4cm]{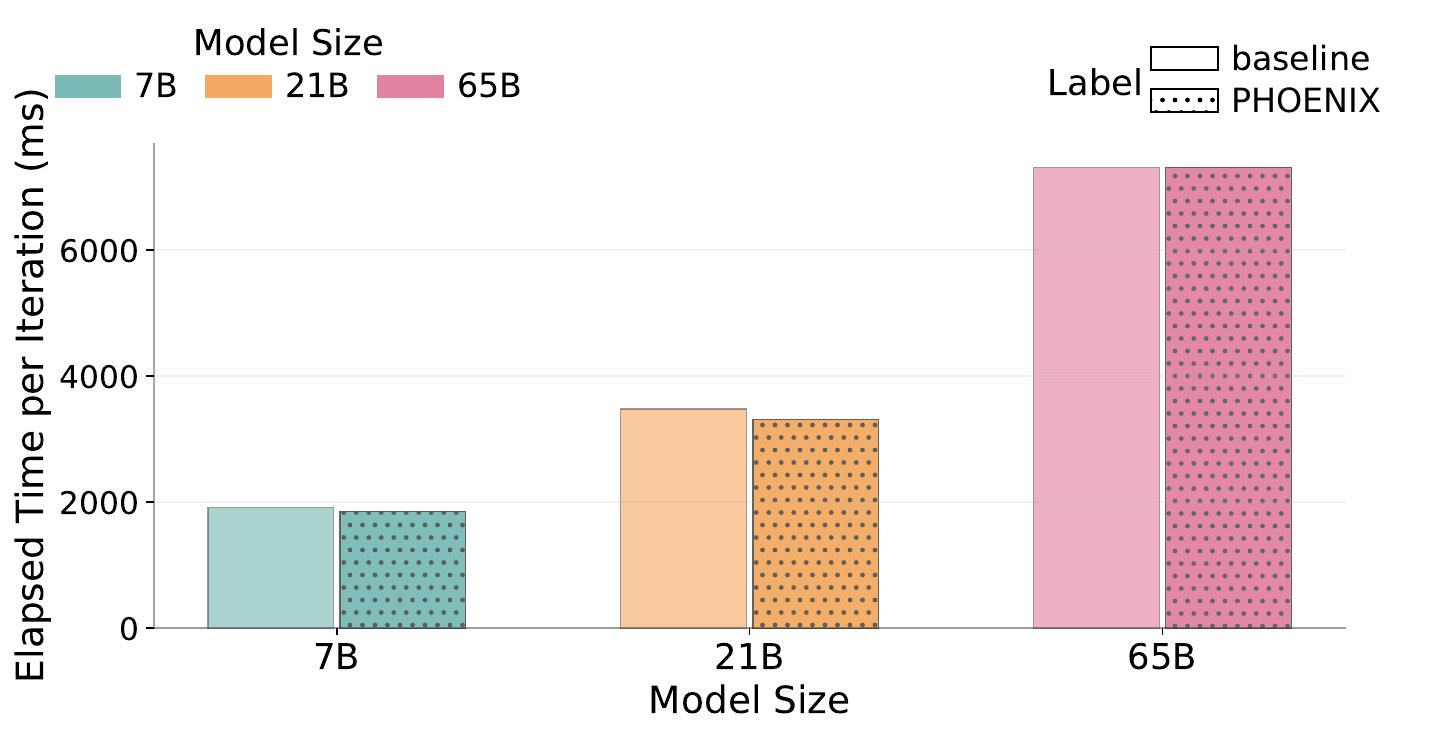}
    \caption{Per-iteration training time for 7B, 21B, and 65B models on Vista.}
    \label{fig:vista}
\end{figure}

To verify that \name{}'s zero-overhead property generalizes across hardware platforms, we further evaluate it with larger models on Vista's Grace--Hopper nodes using 64 nodes. 
\autoref{fig:vista} compares per-iteration training time for 7B, 21B, and 65B models. 
Across all three scales, \name{} closely tracks the baseline: at 7B, \name{} averages 1854.5\,ms versus the baseline's 1912.8\,ms; at 21B, 3311.3\,ms versus 3479.4\,ms; and at 65B, the two are effectively identical (7316.2\,ms vs.\ 7312.0\,ms). 
The small differences in either direction are system noise. 
These results confirm that \name{}'s zero-overhead property holds for models an order of magnitude larger than those tested on Perlmutter, and generalizes across GPU architectures (A100 vs.\ H200) and interconnect fabrics (Slingshot vs.\ InfiniBand).

\subsubsection{\name{} Communication Breakdown}
\label{app:8gpu-overlap}

\begin{figure}
    \centering
    \includegraphics[width=1\linewidth, height=4cm]{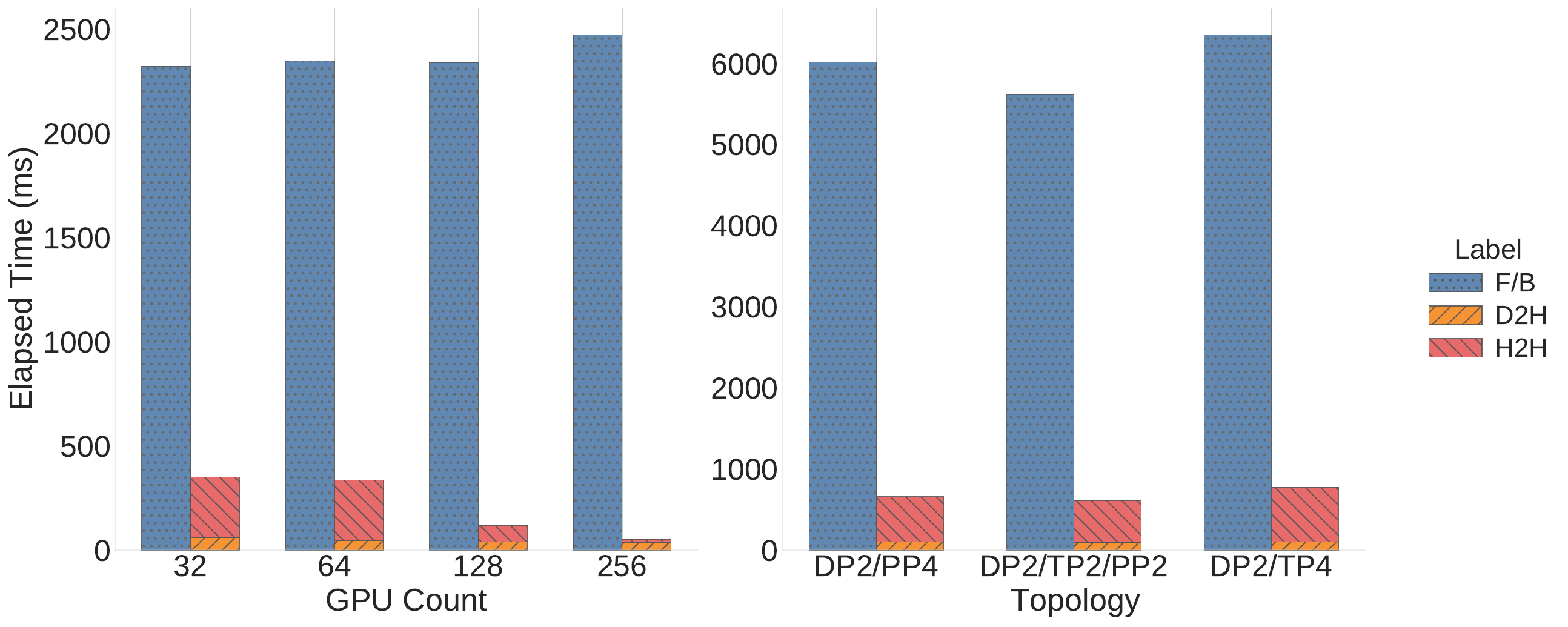}
    \caption{Per-iteration communication breakdown of \name{}.
(\textbf{Left}) Same-topology weak scaling with TP=4 and PP=4, where the x-axis shows the node count and the y-axis shows elapsed time per iteration in milliseconds. 
(\textbf{Right}) 2 nodes runs across three 3D parallel topologies, with the y-axis showing elapsed time per iteration in milliseconds. 
Bars are decomposed into forward/backward computation (F/B), device-to-host transfer (D2H), and host-to-host transfer (H2H).}
    \label{fig:appendix_comm_breakdown}
    \vspace{-3ex}
\end{figure}

We first examine the communication breakdown under the same-topology weak-scaling setup used in the main text. \autoref{fig:appendix_comm_breakdown}(Left) shows that as the system scales from 32 to 256 GPUs with TP=4 and PP=4 fixed, forward/backward computation remains the dominant component of iteration time, while both D2H and H2H offload costs decrease with scale. This behavior is expected because increasing the data-parallel degree reduces the amount of optimizer state that must be offloaded per rank. Consequently, the offload path occupies only a small fraction of the available compute window, making it straightforward to overlap communication with ongoing computation. This explains why \name{} introduces negligible checkpoint overhead in weak-scaling.

We next consider a more communication-intensive setting: 2-node (8-GPU) runs across several 3D parallel topologies. 
In this setting, each rank carries a larger checkpoint payload than in the larger-scale weak-scaling experiments, making overlap more challenging. 
\autoref{fig:appendix_comm_breakdown}(Right) shows that although D2H and H2H costs are higher in absolute terms, they still remain well below the 5.6--6.4\,s forward/backward compute window across all tested topologies. 
Thus, even in this more demanding small-scale case, the offload path still has sufficient room to be overlapped with computation. 
Together, these results show that \name{} benefits naturally from weak scaling, but remains effective even when the per-rank offload volume is higher.

\begin{figure}[t]
    \centering
    \includegraphics[width=1\linewidth, height=5cm]{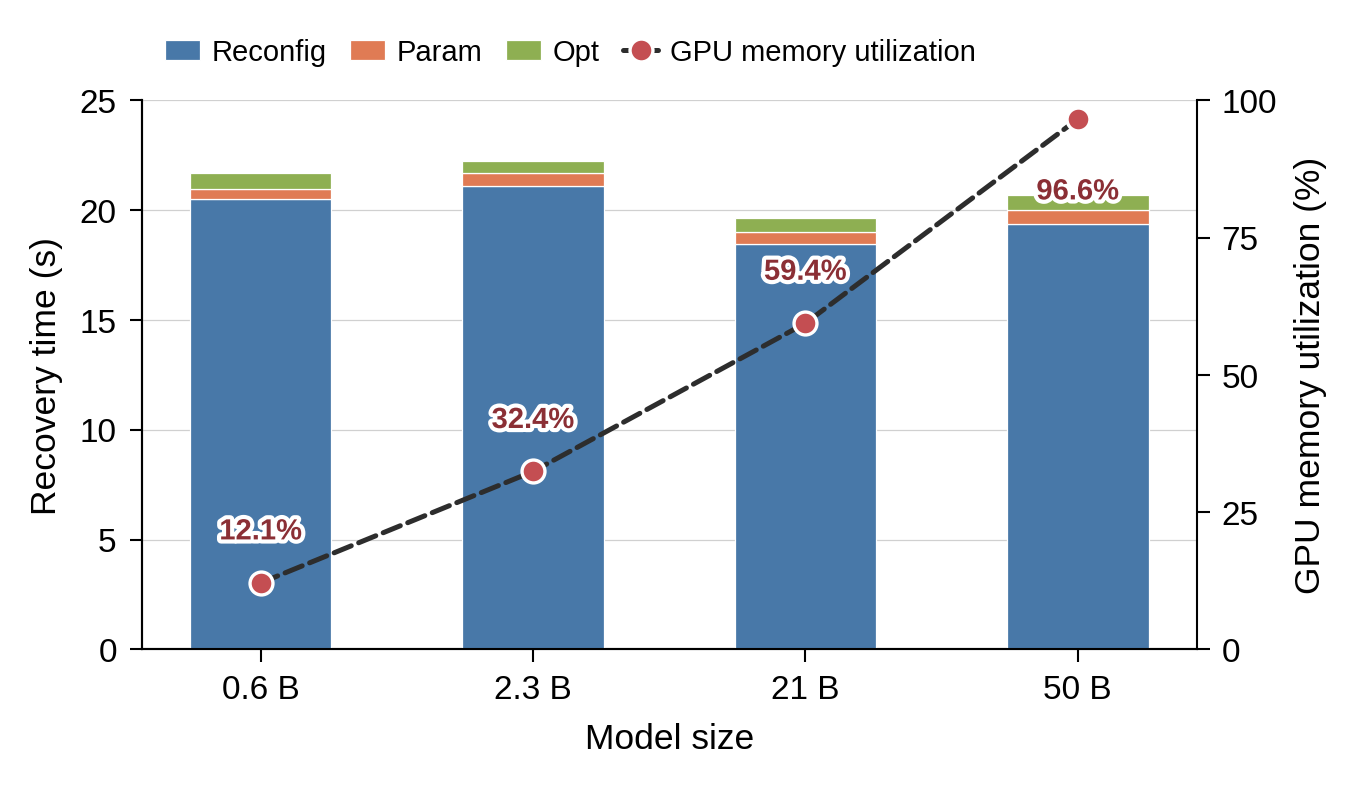}
    \caption{Breakdown of recovery latency across model scales with measured GPU memory utilization on 128 GPUs.}
    \label{fig:recovery_time_gpu_utilization_model_size}
    \vspace{-2ex}
\end{figure}

\begin{figure}[t]
    \centering
    \includegraphics[width=1\linewidth]{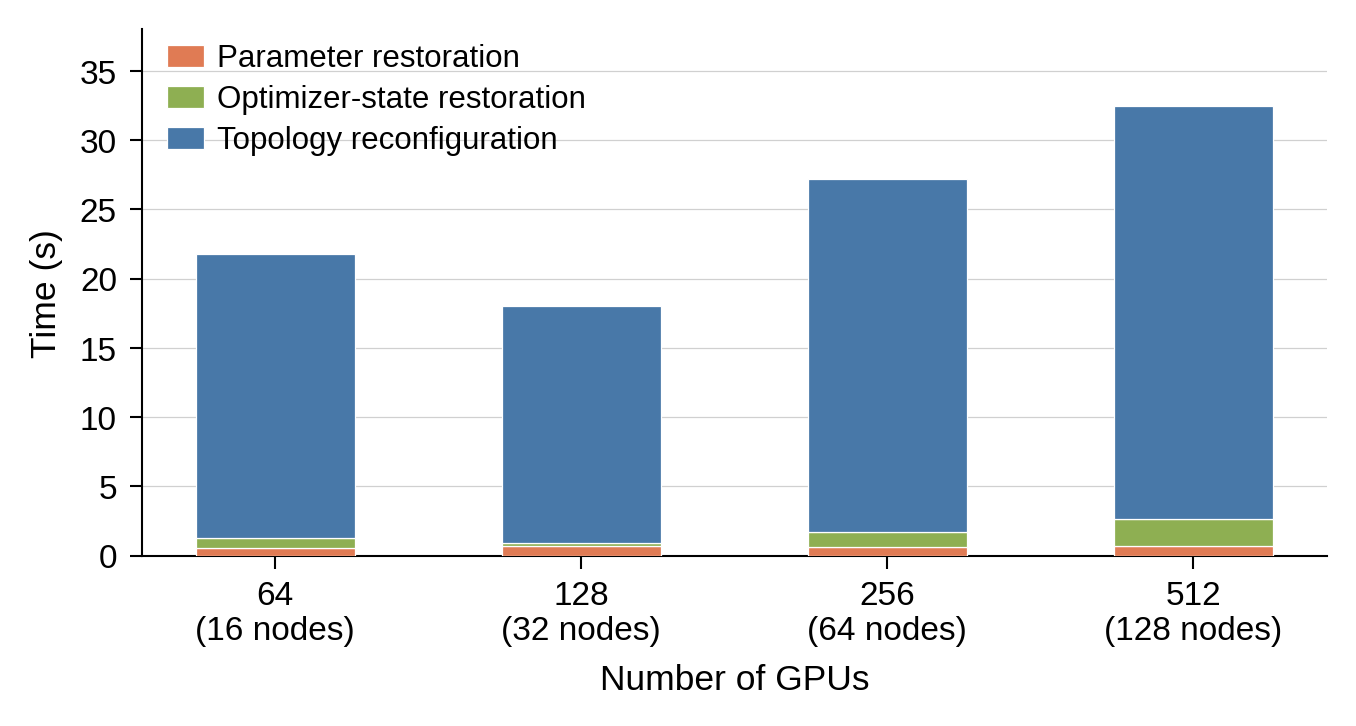}
    \caption{Breakdown of recovery latency and end-to-end recovery time across cluster scales for GPT-7B model under fixed high per-GPU memory utilization.}
    \label{fig:recovery_time_gpu_utilization}
    \vspace{-3ex}
\end{figure}

\subsubsection{Hot-swapping Recovery Latency across Scales}

We evaluate the end-to-end recovery latency of \name{} across both model and system scales.
Our goal is to understand how hot-swapping latency depends on model size, cluster size, and memory pressure.

We first vary model sizes on a fixed 128-GPU cluster with a fixed parallel layout.
As shown in \autoref{fig:recovery_time_gpu_utilization_model_size}, recovery latency remains stable across models ranging from 0.6B to 50B parameters, with per-GPU memory utilization increasing from 12.1\% to over 96.6\%.
This demonstrates that recovery time is independent of model scale and memory footprint.

We then scale the cluster size in a weak-scaling setup, where per-GPU workload and memory usage are held constant.
As shown in \autoref{fig:recovery_time_gpu_utilization}, recovery latency increases from 17.1\,s to 29.8\,s as the system scales from 32 to 512 GPUs. These results show that \name{}'s recovery latency is independent of model size and memory pressure, and grows only with system-level communication reconfiguration
This increase closely follows the growth of topology reconfiguration time, while parameter and optimizer-state restoration remain nearly constant across scales.
A breakdown of the recovery path shows that communicator reconstruction (e.g., NCCL rebuild) dominates recovery latency, whereas state restoration contributes only a small and stable portion.

In comparison, restarting from checkpoint in our experiments takes on average 150\,s, including distributed initialization, checkpoint loading, and process synchronization, while \name{} completes recovery via hot-swapping in under 40\,s.
This corresponds to a reduction of over 3.7$\times$ in recovery latency.
Notably, communicator reconstruction corresponds to a subset of the initialization steps required in checkpoint-restart, where restarting the training job requires full \texttt{torch.distributed} initialization and additional overheads.
In contrast, \name{} limits recovery to communication reconfiguration without full re-initialization, resulting in a more lightweight recovery path.
As system scale increases and failures become more frequent, the relative advantage of \name{} is expected to further increase.

\subsection{Real-World Benefit Quantification}

We quantify the impact of \name{} under realistic failure conditions by combining measured recovery costs from a real training run with failure events generated from empirically observed failure rates.
We use the failure rate reported in Meta's large-scale training study~\cite{salpekar2026training} to model the occurrence of failures, and apply Monte Carlo sampling to generate failure times for our experiment.

We consider a 2.3B training job on 64 nodes for 100K steps, with persistent checkpoints taken every 5K steps (8.9\,s per checkpoint), corresponding to a total runtime of 2.41 days (2.08\,s per step).
Based on the sampled failure process, two failure events occur at 18 and 2366 steps after the most recent checkpoint, respectively.

To resume to the original training progress, \name{} requires 32.6\,s and 29.8\,s for the two failures, respectively.
In contrast, checkpoint-restart requires 150\,s per recovery in addition to replaying lost work since the last checkpoint.
Overall, \name{} reduces the failure-induced recovery time to \textbf{5.3\%} of checkpoint-restart, corresponding to a \textbf{18.8$\times$} reduction in time to resume training progress.
As system scale increases and failures become more frequent, the relative advantage of \name{} becomes more pronounced, since it avoids both repeated checkpoint loading and full-job restarts.

%% file: 08conclusion.tex
\section{Conclusion and Future Work}

In this paper, we present \name{}, a fault-tolerant training system for large-scale LLMs that enables recovery via online hot-swapping rather than full job restarts. 
By continuously replicating optimizer-state shards in memory and reconstructing distributed communication with a spare node, \name{} transforms failure recovery into an online topology repair problem.
We show that \name{} enables per-iteration checkpointing without introducing overhead on the training critical path, making fine-grained state protection practical at scale. Our evaluation further demonstrates that hot-swapping achieves efficient and scalable recovery: recovery latency grows only gradually with cluster size and is dominated by communicator reconstruction, while the cost of state restoration remains nearly constant.
Together, these results indicate that \name{} decouples recovery performance from model size and avoids the replay and restart costs inherent in checkpoint-based approaches. This makes \name{} a practical and scalable solution for fault-tolerant LLM training in large-scale systems.

Future work includes improving control plane fault tolerance, extending \name{} to additional training frameworks, and validating its performance at larger scales.

%% file: refs.bib
@inproceedings{10.1145/2063384.2063427,
author = {Bautista-Gomez, Leonardo and Tsuboi, Seiji and Komatitsch, Dimitri and Cappello, Franck and Maruyama, Naoya and Matsuoka, Satoshi},
title = {FTI: high performance fault tolerance interface for hybrid systems},
year = {2011},
isbn = {9781450307710},
publisher = {Association for Computing Machinery},
address = {New York, NY, USA},
url = {https://doi.org/10.1145/2063384.2063427},
doi = {10.1145/2063384.2063427},
booktitle = {Proceedings of 2011 International Conference for High Performance Computing, Networking, Storage and Analysis},
articleno = {32},
numpages = {32},
location = {Seattle, Washington},
series = {SC '11}
}

@inproceedings{10.5555/3767955.3767985,
author = {Wan, Borui and Han, Mingji and Sheng, Yiyao and Peng, Yanghua and Lin, Haibin and Zhang, Mofan and Lai, Zhichao and Yu, Menghan and Zhang, Junda and Song, Zuquan and Liu, Xin and Wu, Chuan},
title = {ByteCheckpoint: a unified checkpointing system for large foundation model development},
year = {2025},
isbn = {978-1-939133-46-5},
publisher = {USENIX Association},
address = {USA},
booktitle = {Proceedings of the 22nd USENIX Symposium on Networked Systems Design and Implementation},
articleno = {30},
numpages = {20},
location = {Philadelphia, PA, USA},
series = {NSDI '25},
url={https://arxiv.org/abs/2407.20143}
}

@misc{salpekar2026training,
      title={Training {LLM}s with Fault {Tolerant} {HSDP} on 100,000 {GPU}s}, 
      author={Omkar Salpekar and Rohan Varma and Kenny Yu and Vladimir Ivanov and Yang Wang and Ahmed Sharif and Min Si and Shawn Xu and Feng Tian and Shengbao Zheng and Tristan Rice and Ankush Garg and Shangfu Peng and Shreyas Siravara and Wenyin Fu and Rodrigo de Castro and Adithya Gangidi and Andrey Obraztsov and Sharan Narang and Sergey Edunov and Maxim Naumov and Chunqiang Tang and Mathew Oldham},
      year={2026},
      eprint={2602.00277},
      archivePrefix={arXiv},
      primaryClass={cs.DC},
      url={https://arxiv.org/abs/2602.00277}, 
}

@misc{wu2023transom,
      title={TRANSOM: An {Efficient} {Fault}-{Tolerant} {System} for {Training} {LLM}s}, 
      author={Baodong Wu and Lei Xia and Qingping Li and Kangyu Li and Xu Chen and Yongqiang Guo and Tieyao Xiang and Yuheng Chen and Shigang Li},
      year={2023},
      eprint={2310.10046},
      archivePrefix={arXiv},
      primaryClass={cs.DC},
      url={https://arxiv.org/abs/2310.10046}, 
      doi = {10.48550/arXiv.2310.10046}
}

@inproceedings{10.5555/3768039.3768129,
author = {Lian, Xinyu and Jacobs, Sam Ade and Kurilenko, Lev and Tanaka, Masahiro and Bekman, Stas and Ruwase, Olatunji and Zhang, Minjia},
title = {Universal checkpointing: a flexible and efficient distributed checkpointing system for large-scale DNN training with reconfigurable parallelism},
year = {2025},
isbn = {978-1-939133-48-9},
publisher = {USENIX Association},
address = {USA},
booktitle = {Proceedings of the 2025 USENIX Conference on Usenix Annual Technical Conference},
articleno = {90},
numpages = {16},
location = {Boston, MA, USA},
series = {USENIX ATC '25},
doi = {10.48550/arXiv.2406.18820}
}

@inproceedings {264850,
author = {Jayashree Mohan and Amar Phanishayee and Vijay Chidambaram},
title = {{CheckFreq}: Frequent, {Fine-Grained} {DNN} Checkpointing},
booktitle = {19th USENIX Conference on File and Storage Technologies (FAST 21)},
year = {2021},
isbn = {978-1-939133-20-5},
pages = {203--216},
url = {https://www.usenix.org/conference/fast21/presentation/mohan},
publisher = {USENIX Association},
month = feb
}

@inproceedings{wang2023gemini,
  title={Gemini: Fast failure recovery in distributed training with in-memory checkpoints},
  author={Wang, Zhuang and Jia, Zhen and Zheng, Shuai and Zhang, Zhen and Fu, Xinwei and Ng, TS Eugene and Wang, Yida},
  booktitle={Proceedings of the 29th Symposium on Operating Systems Principles},
  pages={364--381},
  year={2023},
  doi = {10.1145/3600006.3613145}
}

@misc{maeng2021understanding,
      title={CPR: Understanding and Improving Failure Tolerant Training for Deep Learning Recommendation with Partial Recovery}, 
      author={Kiwan Maeng and Shivam Bharuka and Isabel Gao and Mark C. Jeffrey and Vikram Saraph and Bor-Yiing Su and Caroline Trippel and Jiyan Yang and Mike Rabbat and Brandon Lucia and Carole-Jean Wu},
      year={2020},
      eprint={2011.02999},
      archivePrefix={arXiv},
      primaryClass={cs.LG},
      url={https://arxiv.org/abs/2011.02999}, 
}

@misc{jeon2019analysis,
      title={Analysis of Large-Scale Multi-Tenant GPU Clusters for DNN Training Workloads}, 
      author={Myeongjae Jeon and Shivaram Venkataraman and Amar Phanishayee and Junjie Qian and Wencong Xiao and Fan Yang},
      year={2019},
      eprint={1901.05758},
      archivePrefix={arXiv},
      primaryClass={cs.DC},
      url={https://arxiv.org/abs/1901.05758}, 
}

@inproceedings{li2023easyscale,
  title={EasyScale: Elastic Training with Consistent Accuracy and Improved Utilization on GPUs},
  author={Li, Mingzhen and Xiao, Wencong and Yang, Hailong and Sun, Biao and Zhao, Hanyu and Ren, Shiru and Luan, Zhongzhi and Jia, Xianyan and Liu, Yi and Li, Yong and Lin, Wei and Qian, Depei},
  booktitle={Proceedings of the International Conference for High Performance Computing, Networking, Storage and Analysis},
  pages={1--14},
  year={2023},
  doi={10.1145/3581784.3607054}
}

@misc{wang2024dlroverrmresourceoptimizationdeep,
      title={DLRover-RM: Resource Optimization for Deep Recommendation Models Training in the Cloud}, 
      author={Qinlong Wang and Tingfeng Lan and Yinghao Tang and Ziling Huang and Yiheng Du and Haitao Zhang and Jian Sha and Hui Lu and Yuanchun Zhou and Ke Zhang and Mingjie Tang},
      year={2024},
      eprint={2304.01468},
      archivePrefix={arXiv},
      primaryClass={cs.DC},
      url={https://arxiv.org/abs/2304.01468}, 
}

@inproceedings{athlur2022varuna,
  title={Varuna: scalable, low-cost training of massive deep learning models},
  author={Athlur, Sanjith and Saran, Nitika and Sivathanu, Muthian and Ramjee, Ramachandran and Kwatra, Nipun},
  booktitle={Proceedings of the Seventeenth European Conference on Computer Systems},
  pages={472--487},
  year={2022},
  url={https://api.semanticscholar.org/CorpusID:243847496}
}

@inproceedings{thorpe2023bamboo,
  title={Bamboo: Making preemptible instances resilient for affordable training of large {DNNs}},
  author={Thorpe, John and Zhao, Pengzhan and Eyolfson, Jonathan and Qiao, Yifan and Jia, Zhihao and Zhang, Minjia and Netravali, Ravi and Xu, Guoqing Harry},
  booktitle={20th USENIX Symposium on Networked Systems Design and Implementation (NSDI 23)},
  pages={497--513},
  year={2023},
  url={https://arxiv.org/abs/2204.12013}
}

@inproceedings{jang2023oobleck,
  title={Oobleck: Resilient distributed training of large models using pipeline templates},
  author={Jang, Insu and Yang, Zhenning and Zhang, Zhen and Jin, Xin and Chowdhury, Mosharaf},
  booktitle={Proceedings of the 29th Symposium on Operating Systems Principles},
  pages={382--395},
  year={2023},
  DOI={10.1145/3600006.3613152}
}

@inproceedings{duan2024parcae,
  title={Parcae: Proactive,{Liveput-Optimized} {DNN} {Training} on {Preemptible Instances}},
  author={Duan, Jiangfei and Song, Ziang and Miao, Xupeng and Xi, Xiaoli and Lin, Dahua and Xu, Harry and Zhang, Minjia and Jia, Zhihao},
  booktitle={21st USENIX Symposium on Networked Systems Design and Implementation (NSDI 24)},
  pages={1121--1139},
  year={2024},
  url={https://arxiv.org/abs/2403.14097}
}

@article{gandhi2024slipstream,
  title={SlipStream: Adapting Pipelines for Distributed Training of Large DNNs Amid Failures},
  author={Gandhi, Swapnil and Zhao, Mark and Skiadopoulos, Athinagoras and Kozyrakis, Christos},
  journal={arXiv preprint arXiv:2405.14009},
  year={2024},
  doi = {10.48550/arXiv.2405.14009}
}

@inproceedings{mohan2021checkfreq,
  title={{CheckFreq}: {Frequent},{Fine-Grained} {DNN} {Checkpointing}},
  author={Mohan, Jayashree and Phanishayee, Amar and Chidambaram, Vijay},
  booktitle={19th USENIX Conference on File and Storage Technologies (FAST 21)},
  pages={203--216},
  year={2021},
  url = {https://par.nsf.gov/biblio/10286595}
}

@article{wang2024fastpersist,
  title={{FastPersist}: {Accelerating} {Model} {Checkpointing} in {Deep} {Learning}},
  author={Wang, Guanhua and Ruwase, Olatunji and Xie, Bing and He, Yuxiong},
  journal={arXiv preprint arXiv:2406.13768},
  year={2024},
  url={https://arxiv.org/abs/2406.13768}
}

@inproceedings{gupta2024just,
  title={Just-In-Time Checkpointing: Low Cost Error Recovery from Deep Learning Training Failures},
  author={Gupta, Tanmaey and Krishnan, Sanjeev and Kumar, Rituraj and Vijeev, Abhishek and Gulavani, Bhargav and Kwatra, Nipun and Ramjee, Ramachandran and Sivathanu, Muthian},
  booktitle={Proceedings of the Nineteenth European Conference on Computer Systems},
  pages={1110--1125},
  year={2024},
  doi = {10.1145/3627703.3650085}
}

@article{zhang2023efficient,
  title={Efficient Fault Tolerance for Recommendation Model Training via Erasure Coding},
  author={Zhang, Tianyu and Liu, Kaige and Kosaian, Jack and Yang, Juncheng and Vinayak, Rashmi},
  journal={Proceedings of the VLDB Endowment},
  volume={16},
  number={11},
  pages={3137--3150},
  year={2023},
  publisher={VLDB Endowment},
  doi = {10.14778/3611479.3611514}
}

@inproceedings{eisenman2022check,
  title={{Check-N-Run}: A checkpointing system for training deep learning recommendation models},
  author={Eisenman, Assaf and Matam, Kiran Kumar and Ingram, Steven and Mudigere, Dheevatsa and Krishnamoorthi, Raghuraman and Nair, Krishnakumar and Smelyanskiy, Misha and Annavaram, Murali},
  booktitle={19th USENIX Symposium on Networked Systems Design and Implementation (NSDI 22)},
  pages={929--943},
  year={2022},
  url = {https://www.usenix.org/conference/nsdi22/presentation/eisenman}
}

@inproceedings{he2023understanding,
  title={Understanding and mitigating hardware failures in deep learning training systems},
  author={He, Yi and Hutton, Mike and Chan, Steven and De Gruijl, Robert and Govindaraju, Rama and Patil, Nishant and Li, Yanjing},
  booktitle={Proceedings of the 50th Annual International Symposium on Computer Architecture},
  pages={1--16},
  year={2023},
  doi = {10.1145/3579371.3589105}
}

@inproceedings{zhang2019quantifying,
  title={Quantifying the impact of memory errors in deep learning},
  author={Zhang, Zhao and Huang, Lei and Huang, Ruizhu and Xu, Weijia and Katz, Daniel S},
  booktitle={2019 IEEE International Conference on Cluster Computing (CLUSTER)},
  pages={1--12},
  year={2019},
  organization={IEEE},
  doi={10.1109/CLUSTER.2019.8890989}
}

@inproceedings{de2017radiation,
  title={Radiation-induced error criticality in modern HPC parallel accelerators},
  author={De Oliveira, Daniel Alfonso Goncalves and Pilla, Laercio Lima and Hanzich, Mauricio and Fratin, Vinicius and Fernandes, Fernando and Lunardi, Caio and Cela, Jos{\'e} Mar{\'\i}a and Navaux, Philippe Olivier Alexandre and Carro, Luigi and Rech, Paolo},
  booktitle={2017 IEEE International Symposium on High Performance Computer Architecture (HPCA)},
  pages={577--588},
  year={2017},
  organization={IEEE},
  doi={10.1109/HPCA.2017.41}
}

@inproceedings{weng2022mlaas,
  title={{MLaaS} in the wild: Workload analysis and scheduling in {Large-Scale} heterogeneous {GPU} clusters},
  author={Weng, Qizhen and Xiao, Wencong and Yu, Yinghao and Wang, Wei and Wang, Cheng and He, Jian and Li, Yong and Zhang, Liping and Lin, Wei and Ding, Yu},
  booktitle={19th USENIX Symposium on Networked Systems Design and Implementation (NSDI 22)},
  pages={945--960},
  year={2022},
  url = {https://www.usenix.org/conference/nsdi22/presentation/weng}
}

@misc{opt175logbook,
  title={OPT-175B Baselines Logbook},
  author={{MetaSeq}},  note="\url{https://github.com/facebookresearch/metaseq/blob/main/projects/OPT/chronicles/OPT175B_Logbook.pdf}",
  year={2022}
}

@article{sergeev2018horovod,
  Author = {Alexander Sergeev and Mike Del Balso},
  Journal = {arXiv preprint arXiv:1802.05799},
  Title = {Horovod: Fast and easy distributed deep learning in {TensorFlow}},
  Year = {2018},
  url={https://arxiv.org/abs/1802.05799}
}

@article{paszke2019pytorch,
  title={Pytorch: An imperative style, high-performance deep learning library},
  author={Paszke, Adam and Gross, Sam and Massa, Francisco and Lerer, Adam and Bradbury, James and Chanan, Gregory and Killeen, Trevor and Lin, Zeming and Gimelshein, Natalia and Antiga, Luca },
  journal={Advances in neural information processing systems},
  volume={32},
  pages={8026--8037},
  year={2019},
  url={https://arxiv.org/abs/1912.01703}
}

@article{hoffmann2022training,
  title={Training compute-optimal large language models},
  author={Jordan Hoffmann and Sebastian Borgeaud and Arthur Mensch and Elena Buchatskaya and Trevor Cai and Eliza Rutherford and Diego de Las Casas and Lisa Anne Hendricks and Johannes Welbl and Aidan Clark and Tom Hennigan and Eric Noland and Katie Millican and George van den Driessche and Bogdan Damoc and Aurelia Guy and Simon Osindero and Karen Simonyan and Erich Elsen and Jack W. Rae and Oriol Vinyals and Laurent Sifre},
  journal={arXiv preprint arXiv:2203.15556},
  year={2022},
  url={https://arxiv.org/abs/2203.15556}
}

@article{ben2019demystifying,
  title={Demystifying parallel and distributed deep learning: An in-depth concurrency analysis},
  author={Ben-Nun, Tal and Hoefler, Torsten},
  journal={ACM Computing Surveys},
  volume={52},
  number={4},
  pages={1--43},
  year={2019},
  publisher={ACM New York, NY, USA},
  doi={https://doi.org/10.1145/3320060}
}

@article{bommasani2021opportunities,
  title={On the opportunities and risks of foundation models}, 
  author={Rishi Bommasani and Drew A. Hudson and Ehsan Adeli and Russ Altman and others},
  journal={Preprint arXiv:2108.07258},
  year={2021},
  url={https://arxiv.org/abs/2108.07258}
}

@misc{liu2023ring,
      title={Ring Attention with Blockwise Transformers for Near-Infinite Context}, 
      author={Hao Liu and Matei Zaharia and Pieter Abbeel},
      year={2023},
      eprint={2310.01889},
      archivePrefix={arXiv},
      primaryClass={cs.CL},
      url={https://arxiv.org/abs/2310.01889}
}

@article{liu2024blockwise,
  title={Blockwise Parallel Transformers for Large Context Models},
  author={Liu, Hao and Abbeel, Pieter},
  journal={Advances in Neural Information Processing Systems},
  volume={36},
  year={2024},
  doi={10.5555/3666122.3666508}
}

@misc{li2022sequence,
      title={Sequence Parallelism: Long Sequence Training from System Perspective}, 
      author={Shenggui Li and Fuzhao Xue and Chaitanya Baranwal and Yongbin Li and Yang You},
      year={2022},
      eprint={2105.13120},
      archivePrefix={arXiv},
      primaryClass={cs.LG},
      url={https://arxiv.org/abs/2105.13120}
}

@inproceedings{nie2016large,
  author       = {Bin Nie and
                  Devesh Tiwari and
                  Saurabh Gupta and
                  Evgenia Smirni and
                  James H. Rogers},
  title        = {A large-scale study of soft-errors on GPUs in the field},
  booktitle    = {2016 {IEEE} International Symposium on High Performance Computer Architecture,
                  {HPCA} 2016, Barcelona, Spain, March 12-16, 2016},
  pages        = {519--530},
  publisher    = {{IEEE} Computer Society},
  year         = {2016},
  url          = {https://doi.org/10.1109/HPCA.2016.7446091},
  doi          = {10.1109/HPCA.2016.7446091},
  bibsource    = {dblp computer science bibliography, https://dblp.org}
}

@inproceedings{beigi2023systematic,
  title={A Systematic Study of DDR4 DRAM Faults in the Field},
  author={Beigi, Majed Valad and Cao, Yi and Gurumurthi, Sudhanva and Recchia, Charles and Walton, Andrew and Sridharan, Vilas},
  booktitle={2023 IEEE International Symposium on High-Performance Computer Architecture (HPCA)},
  pages={991--1002},
  year={2023},
  organization={IEEE},
  doi={10.1109/HPCA56546.2023.10071066}
}

@INPROCEEDINGS{8416467,
  author={Fratin, Vinícius and Oliveira, Daniel and Lunardi, Caio and Santos, Fernando and Rodrigues, Gennaro and Rech, Paolo},
  booktitle={2018 48th Annual IEEE/IFIP International Conference on Dependable Systems and Networks (DSN)}, 
  title={Code-Dependent and Architecture-Dependent Reliability Behaviors}, 
  year={2018},
  volume={},
  number={},
  pages={13-26},
  doi={10.1109/DSN.2018.00015}}

@inproceedings{killi,
  author    = {Shrikanth Ganapathy and
               John Kalamatianos and
               Bradford M. Beckmann and
               Steven Raasch and
               Lukasz G. Szafaryn},
  title     = {Killi: Runtime Fault Classification to Deploy Low Voltage Caches without
               {MBIST}},
  booktitle = {25th {IEEE} International Symposium on High Performance Computer Architecture,
               {HPCA} 2019, Washington, DC, USA, February 16-20, 2019},
  pages     = {304--316},
  publisher = {{IEEE}},
  year      = {2019},
  url       = {https://doi.org/10.1109/HPCA.2019.00046},
  doi       = {10.1109/HPCA.2019.00046},
  bibsource = {dblp computer science bibliography, https://dblp.org}
}

@article{snir2014addressing,
  title={Addressing failures in exascale computing},
  author={Marc Snir and Robert W Wisniewski and Jacob A Abraham and Sarita V Adve and Saurabh Bagchi and Pavan Balaji and Jim Belak and Pradip Bose and Franck Cappello and Bill Carlson and Andrew A Chien and Paul Coteus and Nathan A DeBardeleben and Pedro C Diniz and Christian Engelmann and Mattan Erez and Saverio Fazzari and Al Geist and Rinku Gupta and Fred Johnson and Sriram Krishnamoorthy and Sven Leyffer and Dean Liberty and Subhasish Mitra and Todd Munson and Rob Schreiber and Jon Stearley and Eric Van Hensbergen},
  journal={The International Journal of High Performance Computing Applications},
  volume={28},
  number={2},
  pages={129--173},
  year={2014},
  publisher={Sage Publications Sage UK: London, England},
  doi = {10.1177/1094342014522573}
}

@article{pinciroli2021lifespan,
  title={Lifespan and Failures of SSDs and HDDs: Similarities, Differences, and Prediction Models},
  author={Pinciroli, Riccardo and Yang, Lishan and Alter, Jacob and Smirni, Evgenia},
  journal={IEEE Transactions on Dependable and Secure Computing},
  year={2021},
  publisher={IEEE},
  doi={10.1109/TDSC.2021.3131571}
}

@inproceedings{maurya2024datastates,
  title={Datastates-llm: Lazy asynchronous checkpointing for large language models},
  author={Maurya, Avinash and Underwood, Robert and Rafique, M Mustafa and Cappello, Franck and Nicolae, Bogdan},
  booktitle={Proceedings of the 33rd international symposium on high-performance parallel and distributed computing},
  pages={227--239},
  year={2024},
  doi = {10.1145/3625549.3658685}
}

@article{megatronlm,
  title={Efficient Large-Scale Language Model Training on GPU Clusters Using Megatron-LM},
  author={Narayanan, Deepak and Shoeybi, Mohammad and Casper, Jared and LeGresley, Patrick and Patwary, Mostofa and Catanzaro, Bryan},
  journal={arXiv preprint arXiv:2104.04473},
  year={2021},
  url={https://arxiv.org/abs/2104.04473}
}

@misc{salpekar2026trainingllmsfaulttolerant,
      title={Training LLMs with Fault Tolerant HSDP on 100,000 GPUs}, 
      author={Omkar Salpekar and Rohan Varma and Kenny Yu and Vladimir Ivanov and Yang Wang and Ahmed Sharif and Min Si and Shawn Xu and Feng Tian and Shengbao Zheng and Tristan Rice and Ankush Garg and Shangfu Peng and Shreyas Siravara and Wenyin Fu and Rodrigo de Castro and Adithya Gangidi and Andrey Obraztsov and Sharan Narang and Sergey Edunov and Maxim Naumov and Chunqiang Tang and Mathew Oldham},
      year={2026},
      eprint={2602.00277},
      archivePrefix={arXiv},
      primaryClass={cs.DC},
      url={https://arxiv.org/abs/2602.00277}, 
}

@article{grattafiori2024llama,
  title={The llama 3 herd of models},
  author={Grattafiori, Aaron and Dubey, Abhimanyu and Jauhri, Abhinav and Pandey, Abhinav and Kadian, Abhishek and Al-Dahle, Ahmad and Letman, Aiesha and Mathur, Akhil and Schelten, Alan and Vaughan, Alex and others},
  journal={arXiv preprint arXiv:2407.21783},
  year={2024},
  url={https://arxiv.org/abs/2407.21783}
}

@inproceedings{10.1145/3581784.3607041,
author = {Karrels, Ed and Huang, Lei and Kan, Yuhong and Arora, Ishank and Wang, Yinzhi and Katz, Daniel S. and Gropp, William and Zhang, Zhao},
title = {Fine-grained Policy-driven I/O Sharing for Burst Buffers},
year = {2023},
isbn = {9798400701092},
publisher = {Association for Computing Machinery},
address = {New York, NY, USA},
url = {https://doi.org/10.1145/3581784.3607041},
doi = {10.1145/3581784.3607041},
booktitle = {Proceedings of the International Conference for High Performance Computing, Networking, Storage and Analysis},
articleno = {95},
numpages = {12},
location = {Denver, CO, USA},
series = {SC '23}
}
